\documentclass[10pt,twocolumn,letterpaper]{article}

\usepackage{amsthm}
\usepackage{float}
\usepackage[pagenumbers]{cvpr} 
\newtheorem{theorem}{Theorem}[section]
\newtheorem{lemma}[theorem]{Lemma}
\newtheorem*{lemma2}{Lemma~\ref{lemma:1}}

\usepackage[dvipsnames]{xcolor}
\usepackage[utf8]{inputenc} 
\usepackage[T1]{fontenc}    
\usepackage{url}            
\usepackage{booktabs}       
\usepackage{amsfonts}       
\usepackage{nicefrac}       
\usepackage{microtype}      
\usepackage{amsmath}
\usepackage{graphicx}
\usepackage{bbm}
\usepackage{multirow}
\usepackage{adjustbox}
\usepackage{algorithm} 
\usepackage{algorithmic}
\usepackage{setspace}
\usepackage[algo2e,ruled,vlined]{algorithm2e} 
\definecolor{cvprblue}{rgb}{0.21,0.49,0.74}
\usepackage{caption}
\usepackage[pagebackref,breaklinks,colorlinks,citecolor=cvprblue]{hyperref}

\def\paperID{246} 
\def\confName{CVPR}
\def\confYear{2024}

\title{ProTeCt: Prompt Tuning for Taxonomic Open Set Classification}

\author{Tz-Ying Wu* \quad Chih-Hui Ho* \quad Nuno Vasconcelos\\
University of California, San Diego\\
{\tt\small \{tzw001, chh279, nvasconcelos\}@ucsd.edu}
}

\begin{document}

\maketitle

\begin{abstract}
  Visual-language foundation models, like CLIP, learn generalized representations that enable zero-shot open-set classification. Few-shot adaptation methods, based on prompt tuning, have  been shown to further improve performance on downstream datasets. However, these methods do not fare well in the \emph{taxonomic open set} (TOS) setting, where the classifier is asked to make prediction from label set across different levels of semantic granularity. Frequently, they infer incorrect labels at coarser taxonomic class levels, even when the inference at the leaf level (original class labels) is correct. 
  To address this problem, we propose a prompt tuning technique that calibrates the hierarchical consistency of model predictions. A set of metrics of hierarchical consistency, the Hierarchical Consistent Accuracy (HCA) and the Mean Treecut Accuracy (MTA), are first proposed to evaluate TOS model performance. A new \emph{Prompt Tuning for Hierarchical Consistency} (ProTeCt) technique is then proposed to calibrate classification across label set granularities. Results show that ProTeCt can be combined with existing prompt tuning methods to significantly improve TOS classification without degrading the leaf level classification performance. The code is available at \url{https://github.com/gina9726/ProTeCt}.
\end{abstract}

\vspace{-5pt}
\section{Introduction} \label{sec:intro} \vspace{-5pt}

Vision-language foundation models (FMs) have opened up new possibilities for image classification. They are large models, trained on large corpora, to learn aligned representations of images and text. 
For example, CLIP~\cite{clip} combines text and image encoders trained with 400M image-text pairs in an open vocabulary fashion, using a contrastive loss~\cite{simclr,Chen2020ImprovedBW,Sohn2016ImprovedDM,Oord2018RepresentationLW}. Zero-shot classification can then proceed by leveraging the feature alignments. Each class name is first converted to a text prompt, e.g., ``a photo of [CLASS]," which is fed to the text encoder. The resulting text feature is then used as the parameter vector of a softmax classifier of image feature vectors. 
Since the training does not emphasize any particular classes, CLIP supports open set classification.
Several works~\cite{coop,cocoop,maple,upt} have shown that classification accuracy can be enhanced by fine-tuning the FM on the few-shot setting (i.e. few examples per class). To adapt the model and maintain image-text alignment, these works augment the FM with a few learnable prompts~\cite{coop,cocoop,upt,maple}.
The model parameters are then frozen and only the prompts are optimized. This process is known as \textbf{prompt tuning} and can outperform zero-shot performance, on the dataset of interest. 

\begin{figure}[t!]
    \centering
    \includegraphics[width=\linewidth]{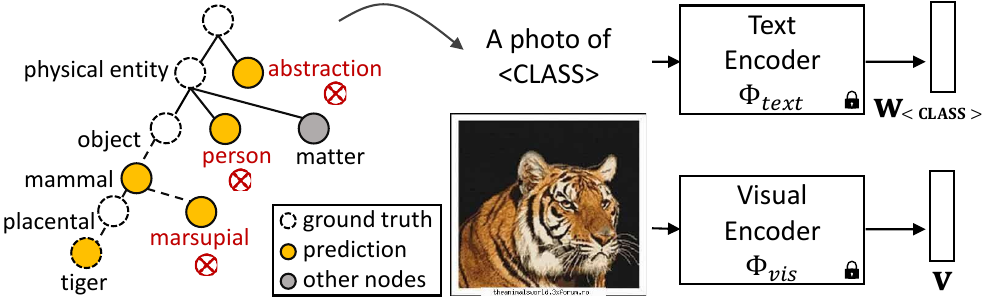}
    \includegraphics[width=0.9\linewidth]{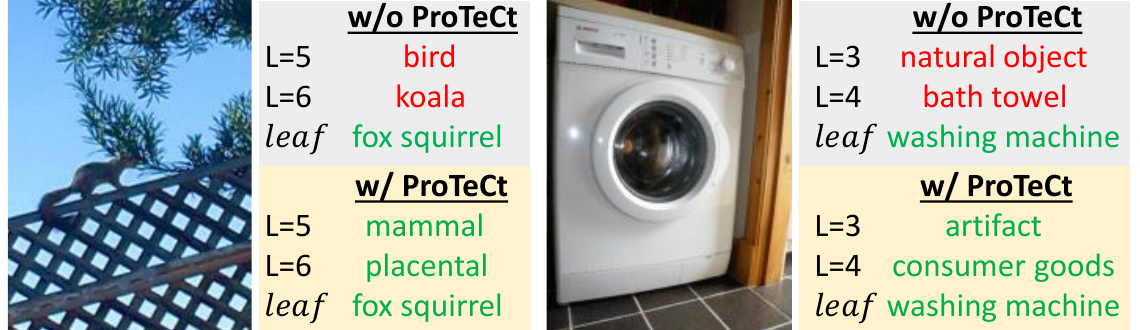}
    \vspace{-5pt}
    \captionof{figure}{(Top) An example of class hierarchy, where CLIP predicts the tiger image as ``person" at the internal hierarchy level. (Bottom) Correct/incorrect model predictions (green/red) of CoOp w/ and w/o ProTeCt on ImageNet variants. $L$ denotes the tree level.}
    \label{fig:teaser}
    \vspace{3pt}
    \setlength{\tabcolsep}{8pt}
    \resizebox{0.75\linewidth}{!}{
    \begin{tabular}{l|ccc}
    \hline
    Method & $Acc_{leaf}$ & HCA & MTA \\
    \hline
    CLIP~\cite{clip} & 68.36 & 3.32 & 48.21\\
    CoOp~\cite{coop} & 71.23 & 2.99 & 46.98 \\
    MaPLe~\cite{maple} & 70.70 &  4.15 & 48.29\\
   \hline
    \end{tabular}
    }
    \vspace{-6pt}
    \captionof{table}{TOS classification performance of CLIP-based classifiers.}
    \label{tab:prelim}
\end{figure}

\vspace{-1pt}
While prompting enables classifiers to be designed for virtually any classes with minimal dataset curation effort, it should not compromise the open set nature and generality of the FM representation. In this work, we consider the setting where ``open set" means the ability to refer to concepts at different levels of granularity. Consider, for example, an educational application in biology. While at grade school level it will teach students to classify animals into (``cat", ``dog", ``lizard"), at the high-school level the {\bf exact same images\/} should be classified into much more detailed classes, e.g. (``iguana", ``anole", ``komodo", etc.) for lizards.  A classifier that classifies an image as a ``komodo" lizard for high schoolers but ``dog" for gradeschoolers is not useful and trustworthy. Advanced biology students should even learn about the taxonomic relations between different species. This requires a representation that supports hierarchical classification~\cite{hd-cnn,Wu20DeepRTC,hier-novel-detect,hier-object-recog}, where the classifier understands the relations between the superclasses and subclasses that compose a class hierarchy, and provides correct predictions {\it across\/} hierarchy levels.

Fig.~\ref{fig:teaser} shows an example hierarchy built from ImageNet~\cite{imagenet} classes, according to the WordNet~\cite{WordNet}. When faced with a tiger image, the classifier should provide a correct prediction under the label sets $\mathcal{Y}_1 =$(``dog'', ``cat", ``{\bf tiger}"), $\mathcal{Y}_2 =$ (``person", ``{\bf animal}", ``insect'') or $\mathcal{Y}_3 =$(``{\bf physical entity}", ``abstraction"), where the correct one is shown in bold. Note that, given a classifier with this property, teachers have the ability to define different classification problems, for many levels of granularity, tailoring the same app to different uses. 
We refer to this setting as \textbf{taxonomic open set} (TOS) classification. In many real-world applications, support for this restricted form of open set classification is much more important than support for unbounded open set classification. In the example above, biology teachers do not really care if the classifier can still discriminate between cars and trucks, or soda cans and wine cups. Hence, these classes are irrelevant to the app developer.

In principle, TOS should be trivially supported by FMs. Even at zero-shot level, it should suffice to specify [CLASS] names at the desired levels of granularity. However, our experiments show that this does not work because the representation of most FMs fails to capture taxonmic relations. 
This is illustrated for CLIP in Fig.~\ref{fig:teaser}. While the model knows that the object is a tiger, it fails to know that it is ``a physical entity" and not an ``abstraction" or that it is a ``placental mammal" and not a ``marsupial," indicating that it only understands class relations locally. It can perform well for the leaf class label set ${\cal Y}_1$ , but cannot reason across abstraction levels, and can thus not support TOS classification.  
To enable TOS, we introduce the  notion of  \textbf{hierarchical consistency}, and a new {\it hierarchical consistency accuracy\/} (HCA) metric,
where classification is defined with respect to a taxonomic tree and its success requires the correct prediction of all superclasses (e.g., mammal, object and physical entity) of each ground truth leave class (e.g., tiger). This is complemented by the notion of \textbf{TOS classification}, where classifiers can have any set of nodes in the class hierarchy as the label set, and a new {\it mean treecut accuracy\/} (MTA) metric, which estimates classification accuracy in this setting. 

Our experiments show that neither CLIP nor existing prompt tuning methods~\cite{coop,cocoop,maple} perform well under the HCA and MTA metrics of the TOS setting. Fig.~\ref{fig:teaser} illustrates the problem and the {\it inconsistent\/} CLIP class predictions (orange dots) across hierarchy levels.
Table~\ref{tab:prelim} compares the standard (leaf) accuracy of the model with  HCA/MTA, under both the zero-shot and two prompt-tuning settings. While the leaf accuracy is quite reasonable, hierarchical consistency is very poor.
To address this problem, we propose a novel prompt-tuning procedure, denoted {\it Prompt Tuning for Hierarchical Consistency\/} (ProTeCt), that explicitly targets the TOS setting. Given a dataset of interest, a class hierarchy is extracted from the associated metadata,
a generic public taxonomy (e.g. WordNet~\cite{WordNet}), or a special purpose taxonomy related to the application (e.g. scientific taxonomies). 
Since FMs support classification with open vocabulary, any node in the hierarchy can be used in the label set of the classifier. Prompts are then learned with the help of two new regularization losses that encourage hierarchical consistency. A {\it dynamic treecut loss\/} (DTL) encourages correct classification at all tree levels by sampling random tree cuts during training. 
A {\it node-centric loss\/} (NCL) contributes additional supervision to each internal tree node to increase classification robustness for all granularities of the hierarchy.

Experiments show that ProTeCt significantly improves the performance of prompt tuning methods, like CoOp~\cite{coop} and MaPLe~\cite{maple}, under TOS setting. Fig.~\ref{fig:teaser} shows the predictions of CoOp at different hierarchy levels before/after adding ProTeCt. Under the HCA/MTA metrics, the improvement can be more than 15/25 points on Cifar100, SUN and ImageNet datasets. Following~\cite{coop,cocoop,maple}, we show that these gains hold for zero-shot domain generalization to several variants of ImageNet~\cite{imagenetv2,imagenet_sketch,imagenet_a,imagenet_r}, showing that hierarchical consistency transfers across datasets. Furthermore, ablations show that ProTeCt can be used with different CLIP architectures, parameter tuning methods and taxonomies.

Overall, this work makes four contributions. First, we introduce the TOS setting, including two novel metrics (HCA and MTA) that evaluate the consistency of hierarchical classification.
Second, we show that neither zero-shot CLIP nor existing prompting methods fare well in this setting. Third, we propose a novel prompt-tuning method for the TOS setting, ProTeCt, which improves hierarchical consistency by combining DTL and NCL losses. The former relies on a dynamic stochastic sampling of label sets involving multiple levels of the hierarchy, while the latter regularizes the classification of every node in the hierarchy. Finally, ProTeCt is shown to outperform vanilla prompt tuning methods on three datasets with different hierarchies. 
Extensive ablations demonstrate that ProTeCt is applicable to different parameter tuning methods, CLIP architectures, taxonomies and the learned hierarchical consistency transfers to unseen datasets from different image domains.

\vspace{-5pt}
\section{Related Work} \vspace{-5pt}

\noindent\textbf{Prompt Tuning of Vision-Language Models.} 
Many large vision-language FMs have been proposed recently~\cite{Zhang2023VisionLanguageMF, Du2022ASO,Wang2023LargescaleMP}. Despite their promising zero-shot performance, several works~\cite{cocoop,coop,vpt,maple} have shown that their few-shot finetuning with a dataset from the target application can further improve performance. Unlike conventional finetuning methods that optimize the entire model, these methods are designed to (a) be parameter efficient and (b) maintain the general purpose feature representation of the FM. Several such tuning methods have been proposed for CLIP~\cite{clip}.
Inspired by prompt tuning techniques from the language literature~\cite{lester-etal-2021-power,li-liang-2021-prefix,liu-etal-2022-p}, CoOp~\cite{coop} inserts learnable prompts at the CLIP text input. CoCoOp~\cite{cocoop} further learns a meta-network to generate an image-conditioned prompt. The idea of connecting image and text prompts is further extended by UPT~\cite{upt} and MaPLe~\cite{maple}. The former learns a unified transformer for generating an image and text prompt, the latter learns a coupling function to generate image prompts from text prompts. LASP~\cite{lasp} proposed a text-to-text cross-entropy loss to regularize the distribution shift when different prompts are used. 
Unlike these works, we investigate the TOS problem, where labels can be drawn from any level in a class taxonomy, and propose prompting techniques to improve hierarchical classification consistency. This is shown to be compatible with several of the above prompt-tuning methods without degrading their leaf classification accuracy.

\noindent\textbf{Hierachical Classifiers.}
Hierarchical classification aims to predict labels at different levels of a class hierarchy. 
Early works \cite{hier-object-recog,large-scale-cate, 42854,share-appearance,hier-prior,hedge-your-bets} date back to the era before deep learning and are not directly applicable to deep learning-based models. Several works~\cite{hd-cnn,taxonomy-regularized,b-cnn, vt-cnn,network-of-experts,Kim18} propose hierarchical classifiers for CNN-based deep models. For example, \cite{taxonomy-regularized,b-cnn,vt-cnn} use additional convolutional modules to learn a hierarchical feature space. It is unclear how these approaches generalize to the recent transformer-based architectures~\cite{vit,swin,swinv2}. Furthermore, prior works~\cite {hd-cnn,taxonomy-regularized,b-cnn, vt-cnn,network-of-experts,Wu20DeepRTC} finetune the entire model, which requires substantial data and computation, especially at the FM scale. In this work, we study the problem of hierarchical consistency for foundational vision-language models (e.g., CLIP). While CLIP-based classifiers~\cite{clip,cocoop,coop} have outstanding zero/few-shot performance, we show that they produce inconsistent predictions for label sets of different granularity and cannot be used in the TOS setting. We propose an efficient prompt tuning method to address this. 

\vspace{-2pt}
\section{Preliminaries}
\vspace{-3pt}

\paragraph{Foundation Models (FMs).}\label{sec:prelim}
Visual-language FMs are composed by a text $\Phi_{text}$ and a visual $\Phi_{vis}$ encoder, which extract features from text and images, respectively. The two encoders are optimized by contrastive training~\cite{Oord2018RepresentationLW,Sohn2016ImprovedDM,Chen2020ImprovedBW,simclr} to create a joint representation for the two modalities. Since the encoders are learned from a large-scale corpus of image-text pairs, the features are general and support various downstream tasks, e.g., image classification~\cite{coop,cocoop,maple,upt} and segmentation~\cite{cris,clipseg}. While in this work we use the CLIP~\cite{clip}, ProTeCt should generalize to other FMs.

\vspace{-7pt}
\paragraph{Image Classification with FMs.}
Given a label set $\mathcal{Y}=\{t_y\}_{y=1}^C$, a zero-shot classifier can be designed in the FM representation space by introducing a weight vector ${\bf w}_y$ per class $y$. These weight vectors  are obtained by simply using the class name $t_y$ (e.g., ``dog") as a text encoder prompt, i.e., ${\bf w}_y=\Phi_{text}(Emb_t(t_y))\in \mathbb{R}^k$, where $Emb_t(\cdot)$ is a word embedding. Given these weight vectors, an image classifier of label set $\mathcal{Y}$ can be implemented by computing class posterior probabilities with  
\begin{equation}
    p(t_y|{\bf x};\mathcal{Y}) = \frac{\exp{(\cos({\bf w}_y, {\bf v})/\tau)}}{\sum_{t_j\in\mathcal Y}\exp{(\cos({\bf w}_j, {\bf v})/\tau)}}~,\label{eq:classifier}
\end{equation}
where $p(t_y|{\bf x};\mathcal{Y})$ is the probability of class label $t_y$ given image ${\bf x}$, ${\bf v}=\Phi_{vis}(Emb_v({\bf x}))\in \mathbb{R}^k$ the visual feature vector, $Emb_v(\cdot)$ an image embedding, $cos(\cdot,\cdot)$ the cosine similarity metric, and $\tau$ a temperature hyperparameter. Classification performance can usually be improved by inferring the classifier parameters ${\bf w}_y$ from multiple text prompts, e.g. by including context words such as a prompt prefix $p=$``a photo of", or $p=$``a drawing of", computing ${\bf w}_y = \Phi_{text}(Emb_t(\{p,t_y\}))$, and ensembling the vectors ${\bf w}_y$ obtained from multiple prompts~\cite{clip, coop}. This, however, requires multiple forward passes through $\Phi_{text}$ during inference and can be undesirable for downstream applications.  

More efficient inference can be achieved with prompt tuning~\cite{coop,cocoop,maple,upt}, which leverages a set of learnable parameters $\{{\bf c}^t_m\}_{m=1}^M$ as context features. These are prepended to each 
class name embedding $Emb_t(t_y)$ as text prompts, to produce the weight vectors ${\bf w}_y=\Phi_{text}(\{{\bf c}^t_1,\dots{\bf c}^t_M, Emb_t(t_y)\})$. Note that each ${\bf c}^t_i$ has the same dimension as the word embedding. Given a training dataset $\mathcal{D}=\{({\bf x}_i, y_i)\}_{i=1}^N$, context features can be end-to-end optimized with the cross-entropy loss 
\begin{align}
    L_{{\mathcal{Y}}}(\mathbf{C}^t) = \frac{1}{N}\sum_{i=1}^N \sum_{t_j\in\mathcal{Y}} -\mathbbm{1}(t_j = t_{y_i})\log p(t_j|{\bf x}_i;\mathcal{Y},\mathbf{C}^t)\label{eq:loss}
\end{align}
for the classifier of (\ref{eq:classifier}), where $\mathbbm{1}(\cdot)$ is the indicator function, and $\mathbf{C}^t$ the matrix of context features. Similarly, learnable prompts $\mathbf{c}^v_i$ can be inserted into the image branch, i.e. ${\bf v}=\Phi_{vis}(\{\mathbf{c}^v_1, \ldots, \mathbf{c}^v_M, Emb_v({\bf x})\})$,
for better visual adaptation~\cite{vpt,upt,maple}. To prevent compromising the generalization of the FM embeddings, the parameters of the two encoders (i.e., $\Phi_{text},\Phi_{vis}$) are frozen in the few-shot setting. In this paper, we consider two prompt tuning variants, CoOp~\cite{coop} and MaPLe~\cite{maple}, the former using learnable prompts in the text branch, and the latter on both branches.

\vspace{-7pt}
\paragraph{Class Taxonomy.}
A class taxonomy ${\cal Y}^{tax}$ organizes classes into a tree where classes of similar semantics are recursively assembled into superclasses, at each graph node (e.g. ``dog" is a superclass of ``Chihuahua" and ``Corgi"). For a tree hierarchy, $\mathcal{T}$, each node $n\in\mathcal{N}$ has a single parent and multiple child nodes $Chd(n)$, where $\mathcal{N}$ is the set of tree nodes.
Given a set of classes $\{t_y\}_{y=1}^C$, a tree hierarchy $\mathcal{T}$ can be built by treating $\{t_y\}_{y=1}^C$ as leaf nodes (where $Chd(t_y)=\emptyset$), i.e., $Leaf(\mathcal{T})=\{t_y\}_{y=1}^C$, and recursively grouping classes in a bottom-up manner until a single root node is created, according to the similarity relationships defined by the taxonomy ${\cal Y}^{tax}$. For example, ImageNet~\cite{imagenet} classes are organized into a tree of 1,000 leaf nodes derived from the WordNet~\cite{WordNet} taxonomy. 
Nodes that are not at the leaves are denoted as internal nodes $\mathcal{N}^{int}=\mathcal{N}\setminus Leaf(\mathcal{T})$.

\vspace{-6pt}
\section{Taxonomic Open Set Classification} \vspace{-6pt}

\noindent{\bf Definition.} A significant advantage of FMs for practical applications is their support for open set classification. Since the classifier of (\ref{eq:classifier}) can be implemented with any class names $t_y$, and the FM is trained with an open vocabulary, it is possible to perform classification for virtually any class. Prompting methods improve the classification of the classes defined by the label set $\cal Y$, but attempt to maintain this generality. However, for most applications ``open set" does not mean the ability to recognize ``any possible word." On the contrary, the whole point of prompt tuning is to enhance the FM for a given application \emph{context}. This context defines what ``open set" truly means for the application. In practice, it frequently means ``all the possible ways" to refer to the classes in ${\cal Y}$. 

One important component of this requirement is the ability to describe classes at different levels of granularity. For example, while user A (a car mechanic) may need to know if an image depicts a ``Fan Clutch Wrench" or a  ``Box-Ended Wrench," user B (a retail store worker) may need to know if the exact same image depicts a ``a mechanic's tool" or a ``plumber's tool." A FM-based classification app should be deployable in both the car garage or the retail store. However, because the app is a tool classification app, the prompted model does not need to be good at recognizing ``lollipops," which are beyond the context of the app. On the other hand, it is undesirable to have to prompt-tune the app for every specific use  or user group. Ideally, it should be possible to prompt tune the FM {\it once,} with respect to the {\it entire\/} class taxonomy ${\cal Y}^{tax}$ of tools. The app can then be deployed to each user base without any retraining, by simply drawing the most suitable class names $t_y$ from ${\cal Y}^{tax}$. We refer to this problem as \textbf{Taxonomic Open Set} (TOS) classification and introduce a formal definition in the remainder of this section.

\vspace{-12pt}
\paragraph{Datasets.} Most existing classification dataset can be used to study the TOS problem, since the very nature of taxonomies is to group objects or concepts into semantic classes of different levels of granularity. Hence, most vision datasets are already labeled taxonomically or adopt classes defined by a public taxonomy,  usually WordNet~\cite{WordNet}. We consider three popular datasets: Cifar100~\cite{cifar100}, SUN~\cite{sun397} and ImageNet~\cite{imagenet}. ImageNet is complemented by the ImageNetv2~\cite{imagenetv2}, ImageNet-S~\cite{imagenet_sketch}, ImageNet-A~\cite{imagenet_a} and ImageNet-R~\cite{imagenet_r} to enable the study of generalization across image domains. For each dataset, the K-shot setting is considered, where K images per class are sampled for training. We consider $K=\{1, 2, 4, 8, 16\}$.

\vspace{-1pt}
\noindent{\bf Label sets.} Given a dataset $\cal D$ and class hierarchy ${\cal Y}^{tax}$ a label set $\cal Y$ is defined at each level of granularity, according the latter. The leaf label set  $\mathcal{Y}_{leaf}$  is defined as the set of classes of $\cal D$ and the class hierarchy $\mathcal{T}$ is build recursively, denoting by  $\mathcal{Y}_n=Chd(n)$ the set of
class labels for the children of node $n$. In our experiments, we adopt the default hierarchy of the SUN dataset and use WordNet~\cite{WordNet} to build the hierarchy for Cifar100 and ImageNet. The resulting class hierarchies are as follows. Cifar100~\cite{cifar100} contains 100 leaf nodes and 48 internal nodes. SUN contains 324 leaf nodes and 19 internal nodes (after pruning 73 leaf classes that have confusing superclasses). ImageNet~\cite{imagenet}, ImageNetv2~\cite{imagenetv2} and ImageNet-S~\cite{imagenet_sketch} share a class hierarchy of 1,000 leaf nodes and 368 internal nodes. ImageNet-A~\cite{imagenet_a} and ImageNet-R~\cite{imagenet_r} only contain 200 subclasses and the corresponding internal nodes from the ImageNet hierarchy.  

\noindent\textbf{Metrics:} 
Given a classifier
\begin{eqnarray}
    \hat{y}({\bf x};\mathcal{Y}) = \arg\max_{t_y\in\mathcal{Y}} p(t_y|{\bf x};\mathcal{Y})
\end{eqnarray}
using a label set $\cal Y$, several metrics are proposed to evaluate TOS performance. 

{\bf Leaf Accuracy} is defined as
\begin{eqnarray}
    Acc_{leaf} = \frac{1}{N} \sum_{i=1}^N \mathbbm{1}[\hat{y}({\bf x}_i;\mathcal{Y}_{leaf}) = t_{y_i}]
    \label{eq:acc_leaf}
\end{eqnarray}
and measures the classification accuracy at the leaves of the taxonomic tree (usually defined as the ``dataset classes"). This enables comparison of hierarchical classifiers to standard, or {\it flat\/}, classifiers which only consider the leaf classes. 

\begin{figure*}[t!]
    \centering
    \includegraphics[width=0.9\linewidth]{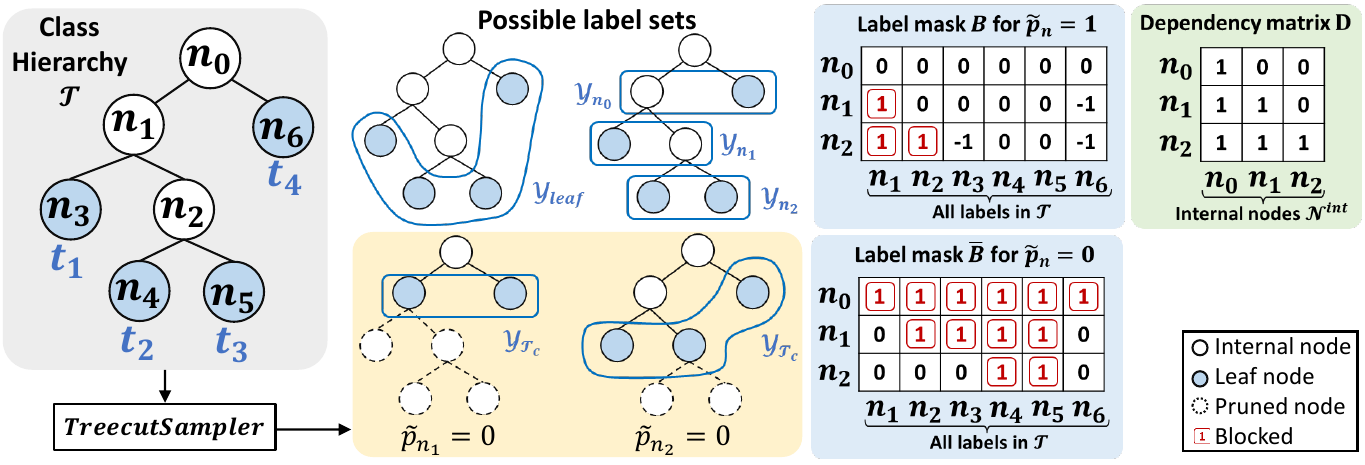}
    \vspace{-10pt}
    \caption{(Left) Multiple possible label sets are available in a class hierarchy. The label set can cover nodes at same level or across different hierarchy levels. (Right) Predefined matrices for efficient treecut sampling used in Algorithm~\ref{algo:treecut_sampler}.}
    \label{fig:tree}
\end{figure*}

{\bf Hierarchical Consistent Accuracy (HCA)} is defined as
\begin{align}
    HCA = \frac{1}{N}\sum_{i=1}^N  & ( \mathbbm{1}[\hat{y}({\bf x}_i;\mathcal{Y}_{leaf}) = t_{y_i}] \nonumber\\ 
    & \prod_{n\in\mathcal{A}(t_{y_i})} \mathbbm{1}[\hat{y}({\bf x}_i;\mathcal{Y}_n)\in \mathcal{A}(t_{y_i})\cup \{t_{y_i}\}]  ) , 
    \label{eq:hca}
\end{align}
where $\mathcal{A}(n)$ denotes all the ancestors of node $n$, and $t_{y_i}$ is the leaf node corresponding to class label $y_i$. While $Acc_{leaf}$ considers successful any correct classification at the leaf level of the tree, the {\it HCA} is stricter. It declares a success only when all the ancestors of the leaf node are correctly classified.
In other words, each sample needs to be classified correctly at each tree level to be viewed as correctly classified under the {\it HCA}. $Acc_{leaf}$ is an upper bound for the {\it HCA}. 

\textbf{Mean Treecut Accuracy (MTA)} estimates the expected accuracy under the TOS classification setting.  It computes the  average accuracy over a set of treecuts $\mathcal{T}_c\in\Omega$,
\begin{eqnarray}
    MTA = \frac{1}{|\Omega|}\sum_{\mathcal{T}_c\in\Omega} \frac{1}{N} \sum_{i=1}^N \mathbbm{1}[\hat{y}({\bf x}_i;\mathcal{Y}_{{\mathcal{T}_c}}) = t_{y_i}]~,\label{eq:mta}
\end{eqnarray}
where $\mathcal{Y}_{\mathcal{T}_c}=Leaf(\mathcal{T}_c)$. However, as shown by the following lemma (see appendix for proof), the set of all possible tree cuts in the hierarchy $\mathcal{T}$ is usually very large. 
\vspace{-3pt}
\begin{lemma}
For a balanced M-ary tree with depth $L$ (root node is excluded and is at depth 0), the number of all valid treecut is $L + \sum_{l=2}^L \sum_{k=1}^{N-1} \frac{N!}{k!(N-k)!} |_{N=M^{l-1}}$.
\label{lemma:1}
\end{lemma}
\vspace{-3pt}
\noindent For example, a tree with $M=2$ and $L=6$ has more than 4 billion treecuts. For a dataset like ImageNet ($L=15$), this number is monumental. Thus, we randomly sampled $|\Omega|=25$ treecuts from $\mathcal{T}$ in all experiments and showed that it is already fairly stable.

\noindent{\bf State-of-the-art.} To test TOS performance of the CLIP with existing prompting techniques, we performed an experiment on ImageNet. Table~\ref{tab:prelim} summarizes the performance of the different methods under the three metrics. Two conclusions are possible. First, the sharp drop from $Acc_{leaf}$ to $HCA$ shows that none of the methods make consistent predictions across the class hierarchy. Second, the low MTAs show that the expected accuracy of TOS classification is dramatically smaller than that of flat classification (leaf classes).

\vspace{-5pt}
\section{Prompt Tuning for Hierarchical Consistency} 
\vspace{-6pt}
To enhance TOS performance of FMs, we propose {\it Prompt Tuning for Hierarchical Consistency\/} (ProTeCt). ProTeCt can be implemented with many existing prompt tuning methods (e.g., CoOp, MaPLe). These methods optimize context prompts using the cross-entropy loss of (\ref{eq:loss}) with leaf label set $\mathcal{Y}_{leaf}$. While this optimizes leaf accuracy $Acc_{leaf}$, it is not robust to label set changes, even for label sets comprised of superclasses of $\mathcal{Y}_{leaf}$. A simple generalization would be to replace (\ref{eq:loss}) with $\mathcal{L}(\mathbf{C}^t) = \sum_{\mathcal{Y}_p \in \mathcal{T}}  L_{{\mathcal{Y}}_p}(\mathbf{C}^t)$, i.e., to consider all the partial label sets $\mathcal{Y}_p$ of the tree $\mathcal{T}$. However, for sizeable taxonomies, this involves a very large number of label sets and is not feasible.
ProTeCt avoids the problem by dynamically sampling label sets from $\mathcal{T}$ during training,  with a combination of two learning objectives, a {\it node-centric loss\/} (NCL) and a {\it dynamic tree-cut loss \/} (DTL).

\vspace{-8pt}
\paragraph{Node-Centric Loss (NCL).} NCL is the aggregate cross-entropy loss of (\ref{eq:loss}) over all node-centric label sets $\mathcal{Y}_n=Chd(n)$ defined by each internal node $n\in\mathcal{N}^{int}$ of the hierarchy, i.e., 
\begin{eqnarray}
    {\cal L}_{NCL}(\mathbf{C}^t) = \frac{1}{|\mathcal{N}^{int}|}\sum_{n\in\mathcal{N}^{int}} L_{\mathcal{Y}_n}(\mathbf{C}^t)~.
\end{eqnarray}
NCL optimization encourages prompts that robustify the classification at different granularities. For example, ``Corgi" should be classified as ``mammal" within the animal label set $\mathcal{Y}_{n_1}=$\{mammal, reptile, bird\}, as a ``dog" in the mammal label set $\mathcal{Y}_{n_2}=$\{dog, cat, elephant, tiger\}, and so forth.

\vspace{-8pt}
\paragraph{Dynamic Treecut Loss (DTL).} While NCL calibrates node classification, guaranteeing consistency within each node, the label sets of TOS classification can also span different sub-trees of the hierarchy, including nodes at different levels, e.g., $\mathcal{Y}=$\{dog, cat, elephant, tiger, reptile, bird\}. DTL seeks to calibrate such label sets, by aggregating the cross-entropy loss of (\ref{eq:loss}) dynamically, i.e., on an example basis, over randomly sampled label sets $\mathcal{Y}_{\mathcal{T}_c}=Leaf(\mathcal{T}_c)$ comprised of the leaves of the tree cuts $\mathcal{T}_c$ (sub-trees) of $\mathcal{T}$.
At each training iteration, a random tree cut $\mathcal{T}_c$ is sampled with the $TreeCutSampler$ procedure of Algorithm~\ref{algo:treecut_sampler}, as illustrated on  the middle of Fig.~\ref{fig:tree}, to define the loss 
\fontsize{8.7pt}{17.4pt}
\begin{eqnarray}
    {\cal L}_{DTL}(\mathbf{C}^t) = L_{\mathcal{Y}_{\mathcal{T}_c}}(\mathbf{C}^t) \quad \mathcal{T}_c \sim TreecutSampler(\mathcal{T}, \beta),
\end{eqnarray}
where $\beta\in[0, 1]$ is a rate of tree dropout. For this, a Bernoulli random variable 
$P_n\sim Bernoulli(\beta)$ of dropout rate $\beta$ is defined for each internal node $n\in\mathcal{N}^{int}\setminus n_0$.
The algorithm descends the tree $\mathcal{T}$, sampling a binary drop-out variable $p_n$ at each node. 
If $p_n=1$, node $n$ is kept in the pruned tree $\mathcal{T}_c$. Otherwise, the sub-tree of  
$\mathcal{T}$ rooted with $n$ is dropped from $\mathcal{T}_c$. The parameter $\beta$ controls the degree of pruning. Larger $\beta$ induces the pruning of more tree nodes, while $\beta=0$ guarantess that $\mathcal{Y}_{\mathcal{T}_c}=\mathcal{Y}_{leaf}$. The root node $n_0$ is excluded, as $p_{n_0}=0$ would imply discarding the whole $\mathcal{T}$. 

\begin{algorithm}[t]\footnotesize
  \setlength{\textfloatsep}{4pt}
  \renewcommand{\linespread}{0.7}
  \SetKwInput{KwInput}{Input}
  \SetKwInput{KwOutput}{Output}
  \KwInput{The tree hierarchy $\mathcal{T}$ of the dataset, tree dropout rate $\beta$}
  \KwOutput{The treecut label set $\mathcal{Y}_{\mathcal{T}_c}$}

  \tcp{sampling ${\bf p}$ for internal nodes; prune the sub-tree rooted at $n$ if $p_n=0$}
  $p_{n_0} \leftarrow 1$ \tcp*{always keep the root node}  
  \For{$n \in \mathcal{N}^{int}\setminus n_0$}{ 
    $p_n \leftarrow Bernoulli(\beta)$\; 
  }
  ${\bf p} \leftarrow (p_{n_1^{int}},...,p_{n_{K}^{int}})$\; \smallskip

  \tcp{correct ${\bf p}$ based on the node dependency}
  ${\bf\tilde p} \leftarrow {\bf p}\otimes\mathbbm{1}[\bf Dp = D1]$ \smallskip

  \tcp{obtain blocked labels with predefined masks and the sampled ${\bf\tilde p}$}
  ${\bf b} \leftarrow min({\bf B}, 0)^T \tilde{\bf p} + \bar{\bf B}^T(\mathbf{1}-\tilde{\bf p})$ \smallskip

  \tcp{gather available (unblocked) labels as the sampled label set}
  $\mathcal{Y}_{\mathcal{T}_c} \leftarrow \{n_j: n_j \in \mathcal{N}\setminus n_0, {\bf b}_j=0\}$ \smallskip
  
  \Return{$\mathcal{Y}_{\mathcal{T}_c}$}\;
  \caption{Treecut Sampler}\label{algo:treecut_sampler}
  \vspace{-0mm}
\end{algorithm}

The $TreeCutSampler$ algorithm is an efficient procedure to sample tree cuts $\mathcal{T}_c$ from $\mathcal{T}$. It starts by sampling a vector ${\bf p}=(p_{n_1^{int}},...,p_{n_{K}^{int}})$, where ${n_i^{int}}$ denotes the $i$-th internal node and $K=|\mathcal{N}^{int}|$, containing pruning flags $p_n$ for all internal nodes $n\in\mathcal{N}^{int}$. The next step is to enforce consistency between these flags, according to the tree structure. If any node in $\mathcal{A}(n)$ is pruned, then node $n$ should be pruned even if $p_n=1$. This is efficiently enforced across all the flags by defining a dependency matrix ${\bf D}\in{\{0, 1\}}^{K\times K}$ where ${\bf D}_{ij}=\mathbbm{1}[n_j^{int}\in \mathcal{A}(n_i^{int})\cup \{n_i^{int}\}]$ indicates whether the $i$-th internal node $n_i^{int}$ is a child of the $j$-th internal node $n_j^{int}$. 
An example is provided on the right of Fig.~\ref{fig:tree} for the tree on the left. The sampled flags are then corrected by computing ${\bf\tilde p}={\bf p}\otimes\mathbbm{1}[\bf Dp = D1]$, where $\bf 1$ is the vector of $K$ ones and $\otimes$ the Hadamard product. Note that both $\mathbf{D}$ and $\mathbf{D1}$ are pre-computed, making the complexity of this step roughly that of one matrix-vector multiplication. 

To identify the leaves of the sampled treecut ($\mathcal{Y}_{\mathcal{T}_c}=Leaf({\mathcal{T}_c})$) efficiently, a mask ${\bf B}\in{\{0, 1, -1\}}^{K\times|\mathcal{N}\setminus \{n_0\}|}$ is defined, where each row corresponds to an internal node, and the columns contain all possible labels in $\mathcal{T}$, i.e., all nodes except the root $n_0$. Entry $B_{ij}$ flags that $n_j$ cannot appear in the sampled label set, given that $n_i\in\mathcal{N}^{int}$ has not been pruned (i.e., $\tilde{p}_{n_i^{int}}=1$), as follows
\fontsize{8.7pt}{17.4pt}
\begin{align}
    B_{ij}= \begin{cases}
            1, ~\text{if}~n_j\in \mathcal{A}(n_i^{int})\cup \{n_i^{int}\} ~~ \textit{\footnotesize ($n_j$ is an ancestor of $n_i^{int}$)}\\
            0, ~\text{if}~n_i^{int}\in \mathcal{A}(n_j) ~~ \textit{\footnotesize ($n_j$ is a descendant of $n_i^{int}$)}\\
           -1, ~\text{otherwise} ~~\textit{\footnotesize ($n_j$ is outside of the sub-tree rooted at $n_i^{int}$)}
        \end{cases}.
\end{align}
Similarly, a matrix $\bar{\bf B}$, of entries $\bar{B}_{ij} = 1 - |B_{ij}|$, is defined to flag that $n_j$ cannot appear in the label set, given  that $n_i\in\mathcal{N}^{int}$ has been pruned, i.e. $\tilde{p}_{n_i^{int}}=0$. A mask of the nodes unavailable to the label set is then computed by accumulating the masks corresponding to the values of $\tilde{\bf p}$,
\begin{eqnarray}
    {\bf b} = min({\bf B}, 0)^T \tilde{\bf p} + \bar{\bf B}^T(\mathbf{1}-\tilde{\bf p})~,
\end{eqnarray}
where the mask in $min({\bf B}, 0)$ is selected if $\tilde{p}_n=1$, and that in $\bar{\bf B}$ if $\tilde{p}_n=0$. Note that $min({\bf B}, 0)$ clips $B_{ij}=-1$ to $0$. The mask ${\bf b}$ can then be used to obtain $\mathcal{Y}_{\mathcal{T}_c}=Leaf({\mathcal{T}_c})=\{n_j: n_j \in \mathcal{N}\setminus n_0, b_j=0\}$.
Fig.~\ref{fig:tree} gives an example. When $\tilde{\bf p}=(\tilde{p}_{n_0},\tilde{p}_{n_1},\tilde{p}_{n_2})=(1, 0, 0)$, then ${\bf b} = min({\bf B}_1, 0) + {\bar{\bf B}}_2 + {\bar{\bf B}}_3 = (0, 1, 1, 2, 2, 0)$, signaling that only $n_1$ and $n_6$ are available to the label set (as $b_1,b_6=0$), resulting in $\mathcal{Y}_{\mathcal{T}_c}=\{n_1, n_6\}$. More detailed examples are given in the appendix.

\vspace{-10pt}
\paragraph{Optimization.}
The overall loss used for prompt tuning is a combination of the two losses
\begin{eqnarray}
    {\cal L}(\mathbf{C}^t)  =  {\cal L}_{DTL}(\mathbf{C}^t)  + \lambda   {\cal L}_{NCL}(\mathbf{C}^t) \label{eq:overallL}
\end{eqnarray}
where $\lambda$ is a hyperparameter.
Note that, like previous prompting approaches, ProTeCt optimizes the learnable prompts $\{{\bf c}_m\}_{m=1}^M$ while keeping the parameters of $\Phi_{text}$, $\Phi_{vis}$ frozen.

\vspace{-3pt}
\section{Experiments}  \vspace{-5pt}
In this section, we discuss experiments for evaluating the effectiveness of ProTeCt. To demonstrate that ProTeCt is a plug-an-play method, it was applied to two SOTA prompt tuning methods: CoOp~\cite{coop} and MaPLe~\cite{maple}. Each experiment is averaged over 3 runs and full tables with error bars are shown in the appendix for brevity. All experiments were conducted on a single Nvidia A10 GPU, using Pytorch~\cite{pytorch}. Please see the appendix for more training details and results. ProTeCt code builds on the publicly available codebases for CoOp and MaPLe and will be released upon publication.

\begin{table*}[t]
    \centering
    
    \adjustbox{max width=\linewidth}{%
    \setlength{\tabcolsep}{4pt}
    \begin{tabular}{|c|cc|cccc|ccc|ccc|}
    \hline
    \multirow{2}{*}{Method} & K- & w/ & \multicolumn{4}{|c|}{Cifar100} & \multicolumn{3}{|c|}{SUN} & \multicolumn{3}{|c|}{ImageNet} \\
    & Shot & ProTeCt &  $Acc_{leaf}$ & HCA & MTA (25)  & MTA (100) & $Acc_{leaf}$ & HCA & MTA (25)  & $Acc_{leaf}$ & HCA & MTA (25) \\
    \hline
    \hline
    \multirow{6}{*}{CoOp} & 16 & &  72.88 & 10.04 & 50.64  & 51.14 & 73.82	& 38.28 & 52.99 & 71.23 & 2.99 & 46.98\\
    & 16 & \checkmark &  72.94 & 56.85 & 87.69 & 87.30 & 74.59 & 62.94 & 83.51 & 69.92 & 37.74 & 88.61\\
    & & &  ({\color{red}+0.06}) & ({\color{red}+46.81}) & ({\color{red}+37.05}) & ({\color{red}+36.16}) &
    ({\color{red}+0.77}) & ({\color{red}+24.66}) & ({\color{red}+30.52}) & ({\color{blue}-1.31}) & ({\color{red}+34.75}) & ({\color{red}+41.63})\\
    \cline{2-13}
    & 1  &  & 65.03 & 7.81 & 41.78 & 44.17 & 63.65 & 33.36 & 51.20 & 63.67 & 1.59 & 40.52\\
    & 1  & \checkmark &  66.88 & 41.01 & 81.64 & 81.01 & 63.79 & 49.62 & 76.25 & 66.11 & 25.79 & 86.14\\
    & & &  ({\color{red}+1.85}) & ({\color{red}+33.2}) & ({\color{red}+39.86}) &  
    ({\color{red}+36.84}) &
    ({\color{red}+0.14}) & ({\color{red}+16.26}) & ({\color{red}+25.05}) & ({\color{red}+2.44}) & ({\color{red}+24.2}) & ({\color{red}+45.62})\\
    \hline
    \hline
    \multirow{6}{*}{MaPLe} & 16 & & 75.01 & 17.54 & 52.21 & 50.82 & 71.86 & 33.25  & 54.29 & 70.70 & 4.15 & 48.29\\
    & 16 & \checkmark & 75.34 & 61.15 & 88.04 & 88.33 & 72.17 & 59.71 & 82.27 & 69.52 & 31.24 & 87.87\\
    & & & ({\color{red}+0.33}) & ({\color{red}+43.61}) & ({\color{red}+35.83}) &  
    ({\color{red}+37.51}) &
    ({\color{red}+0.31}) & ({\color{red}+26.46}) & ({\color{red}+27.98}) &  ({\color{blue}-1.18}) & ({\color{red}+27.09}) &  ({\color{red}+39.58})\\
    \cline{2-13}
    & 1  &  & 68.75 & 4.65 & 50.60 & 54.99 &  63.98 & 25.15 & 50.31 & 68.91 & 2.97 & 48.16\\
    & 1  & \checkmark & 69.33 & 48.10 & 83.36 & 83.78 &  64.29 & 50.45 & 76.73 & 66.16 & 20.44 & 85.18\\
    & & & ({\color{red}+0.58}) & ({\color{red}+43.45}) & ({\color{red}+32.76}) & 
    ({\color{red}+28.79}) &
    ({\color{red}+0.31}) & ({\color{red}+25.30}) & ({\color{red}+26.42}) &  ({\color{blue}-2.75}) & ({\color{red}+17.47}) & ({\color{red}+37.02})\\
    \hline
    \end{tabular}
    }
    \vspace{-8pt}
    \caption{TOS performance w/ and w/o ProTeCt on Cifar100 ($\lambda=0.5$), SUN ($\lambda=0.5$) and ImageNet ($\lambda=1$). $\beta=0.1$ for all datasets.}
    \label{tab:main_table}
\end{table*}

\begin{figure*}
\begin{minipage}[t]{0.29\linewidth}
    \centering
    \vspace{0pt}
    \begin{tabular}{c}
      \includegraphics[width=0.94\linewidth]{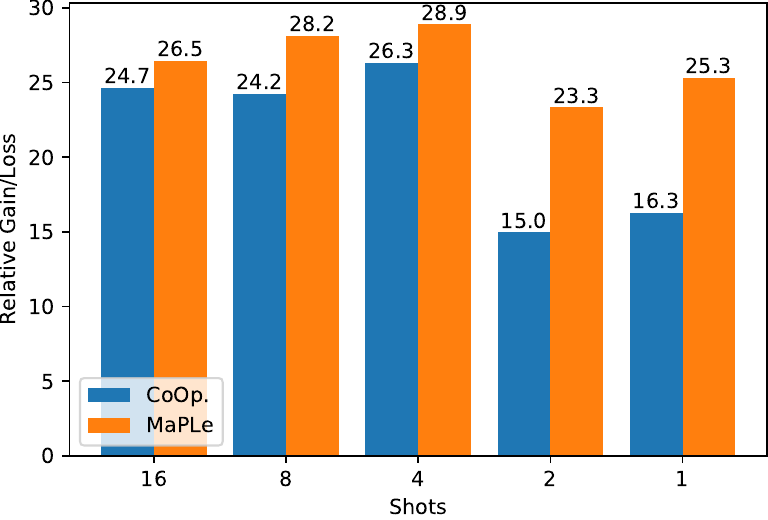} \\
      \includegraphics[width=0.94\linewidth]{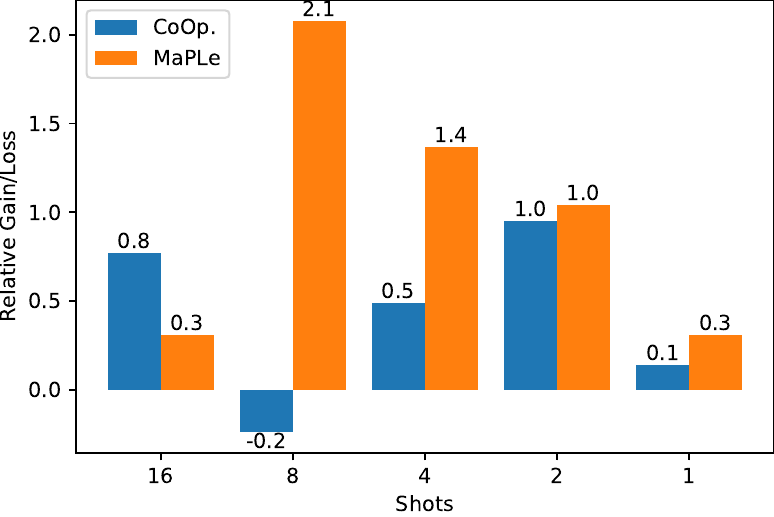} 
    \end{tabular}
    \vspace{-14pt}
    \captionof{figure}{Relative gain/loss after adding ProTeCt to CoOp and MaPle, respectively. (Top) HCA ; (Bottom) $Acc_{leaf}$.}
    \label{fig:rel_gain_all_dataset}
\end{minipage}
\hspace{3pt}
\begin{minipage}[t]{0.7\linewidth}
    \vspace{-0pt}
    \begin{minipage}[t]{\linewidth}
    \resizebox{\linewidth}{!}{
    \begin{tabular}{ccc}
        \includegraphics[width=0.33\linewidth]{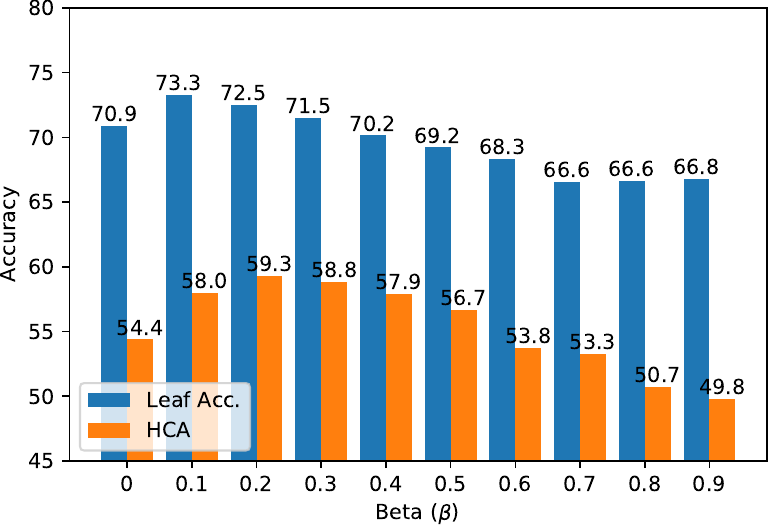} & 
          \includegraphics[width=0.33\linewidth]{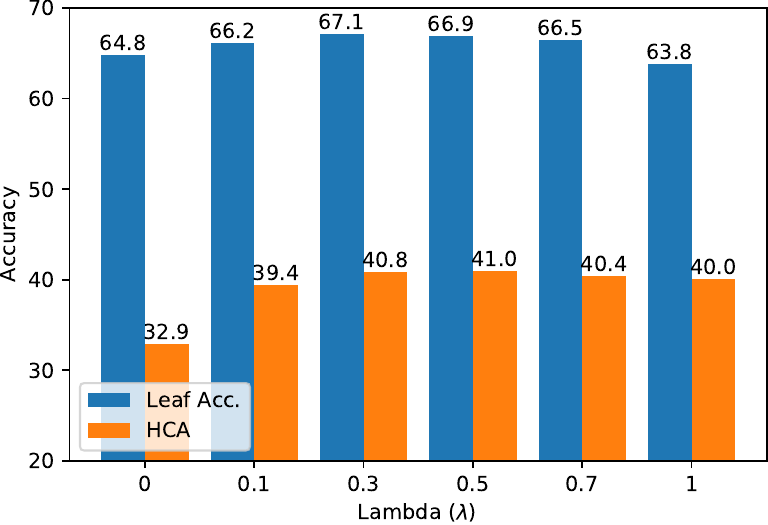}   & 
          \includegraphics[width=0.33\linewidth]{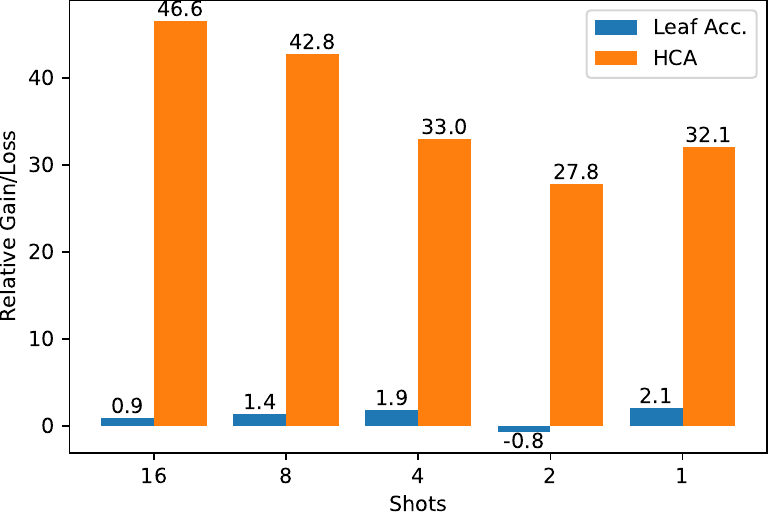} \\
          \vspace{-2pt}
          \footnotesize{(a)} & \footnotesize{(b)} & \footnotesize{(c)}      
        \end{tabular}
        }
        \vspace{-7pt}
        \caption{Ablation of (a) tree dropout rate $\beta$, (b) NCL strength $\lambda$ and (c) CLIP ViT B32 architecture.}
        \label{fig:ablation}
    \end{minipage}
    \begin{minipage}[b]{\linewidth}
    \vspace{10pt}
    \centering
    \resizebox{\linewidth}{!}{
    \begin{tabular}{cccc}     
        \centering
        \includegraphics[width=0.22\linewidth,height=0.22\linewidth]{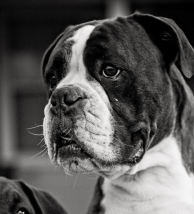} &
        \includegraphics[width=0.22\linewidth,height=0.22\linewidth]{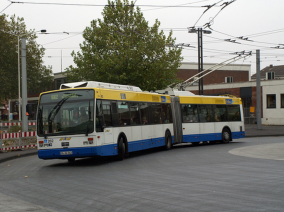} &
        \includegraphics[width=0.22\linewidth,height=0.22\linewidth]{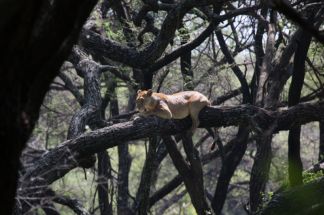} &
        \includegraphics[width=0.22\linewidth,height=0.22\linewidth]{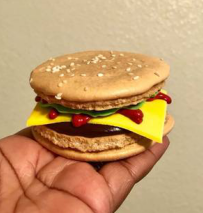} \\
        \footnotesize{(a): [{\color{blue}{Boxer}}, {\color{red}{Person}}] } & \footnotesize{(b):  [{\color{blue}{Trolleybus}}, {\color{red}{Animal}}]  }  & 
        \footnotesize{(c):[{\color{blue}{Lion}}, {\color{red}{Koala}}] } & \footnotesize{(d): [{\color{blue}{Cheeseburger}}, {\color{red}{Ice cream}}]}
    \end{tabular}
    }
    \vspace{-5pt}
    \captionof{figure}{ProTeCt correctly predicts examples from ImageNet (a,b) and its variants (c,d) at all levels. [{\color{blue}{GT}}, {\color{red}{Prediction}}] shows the {\color{blue}{groundtruth}} and {\color{red}{incorrect prediction}} by vanilla prompt tuning.}\label{fig:vis}
    \end{minipage}
    \vfill
\end{minipage}
\end{figure*}

\vspace{-8pt}
\paragraph{Metrics:} $Acc_{leaf}$ of (\ref{eq:acc_leaf}), HCA of (\ref{eq:hca}) and MTA of (\ref{eq:mta}) are considered. MTA uses 5 tree dropout rates ($\beta \in \{0.1, 0.3, 0.5, 0.7, 0.9\}$) to sample treecuts of various granularities. For each $\beta$, $T$ treecuts are sampled without repetition to obtain a total of $5T$ treecuts. MTA($5T$) indicates the result is averaged over these $5T$ treecuts. We ablate $T=5$ and $T=20$ on Cifar100 and use $T=5$ for all datasets by default.

\begin{table*}[t]
    \centering
    \setlength{\tabcolsep}{4pt}
    \adjustbox{max width=\linewidth}{%
    \begin{tabular}{|c|cc|ccc|ccc|ccc|ccc|}
    \hline
    \multirow{2}{*}{Method} & K- & w/ & \multicolumn{3}{|c|}{ImageNetv2~\cite{imagenetv2}} & \multicolumn{3}{|c|}{ImageNet-S~\cite{imagenet_sketch}} & \multicolumn{3}{|c|}{ImageNet-A~\cite{imagenet_a}} & \multicolumn{3}{|c|}{ImageNet-R~\cite{imagenet_r}} \\
    & Shot & ProTeCt &  $Acc_{leaf}$ & HCA & MTA (25) & $Acc_{leaf}$ & HCA & MTA (25)  & $Acc_{leaf}$ & HCA & MTA (25) & $Acc_{leaf}$ & HCA & MTA (25) \\
    \hline
    \hline
    \multirow{6}{*}{CoOp} & 16 & &  64.01 & 2.31 & 43.74 & 47.82 & 1.39 & 38.58 & 50.28 & 2.97 & 52.56 & 75.83 & 18.49 & 64.13\\
     & 16  & \checkmark &   62.60 & 32.84 & 86.66 & 46.80 & 20.73 & 82.60 & 49.08 & 22.45 & 78.21 & 74.94 & 31.18 & 75.59\\
     & & &  {\color{blue}(-1.41)} &  {\color{red}(+30.53)} & {\color{red}(+42.92)} & {\color{blue}(-1.02)} &  {\color{red}(+19.34)} & {\color{red}(+44.02)} &  {\color{blue}(-1.20)} &  {\color{red}(+19.48)} & {\color{red}(+25.65)}&  {\color{blue}(-0.89)} &  {\color{red}(+12.69)}  & {\color{red}(+11.40)}\\
    \cline{2-15}
    & 1  &  & 56.43 & 1.51 & 38.27 & 41.38 & 1.11 & 33.61 & 45.92 & 1.76 & 47.54 & 69.84 & 11.74 & 55.31\\
    & 1  & \checkmark & 60.16 & 22.95 & 84.38 & 44.75 & 13.88 & 80.64 & 48.95 & 20.52 & 76.95 & 74.26 & 27.46 & 76.48\\
    & & &  {\color{red}(+3.73)} &  {\color{red}(+21.44)} & {\color{red}(+46.11)} & {\color{red}(+3.37)} &  {\color{red}(+12.77)} & {\color{red}(+47.03)}& {\color{red}(3.03)} &  {\color{red}(+18.76)} & {\color{red}(+29.41)} & {\color{red}(+4.42)} &  {\color{red}(+15.72)} & {\color{red}(+21.17)}\\
    \hline
    \hline
    \multirow{6}{*}{MaPLe} & 16 &  & 64.15	& 1.97  & 45.93 & 48.97 & 1.58 & 43.37 & 50.61 & 2.31 & 54.88 & 76.61 & 20.67 & 63.06\\
    & 16 & \checkmark & 62.77 & 27.86 & 86.14 & 47.47 & 17.77 & 82.52 & 47.41 & 19.75  & 77.46 & 75.70 & 32.58 & 77.99\\
    & & &  {\color{blue}(-1.38)} &  {\color{red}(+25.89)} & {\color{red}(+40.21)} & {\color{blue}(-1.50)} &  {\color{red}(+16.19)} & {\color{red}(+39.15)} & {\color{blue}(-3.20)} &  {\color{red}(+17.44)} & {\color{red}(+22.58)} & {\color{red}(-0.91)} &  {\color{red}(+11.91}) & {\color{red}(+14.93)}\\
    \cline{2-15}
    & 1  &  & 61.78	& 2.18 & 45.50 & 46.79 & 1.70 & 45.26 & 47.55 & 3.52 & 55.48 & 74.55 & 18.85 & 62.48\\
    & 1  & \checkmark & 59.14 & 17.89 & 83.27 & 44.92 & 11.24 & 79.94 & 47.15 & 16.03 & 76.81 & 74.60 & 25.20 & 75.72\\
    & & &  {\color{blue}(-2.64)} &  {\color{red}(+15.71)} & {\color{red}(+37.77)} &  {\color{blue}(-1.87)} &  {\color{red}(+9.54)} & {\color{red}(+34.68)} & {\color{blue}(-0.40)} &  {\color{red}(+12.51)} & {\color{red}(+21.33)}  &  {\color{red}(+0.05)} &  {\color{red}(+6.35)} & {\color{red}(+13.24)}\\
    \hline
    \end{tabular}
    }
    \vspace{-7pt}
    \caption{The gain of hierarchical consistency after adding ProTeCt generalizes across datasets in unseen domains. All methods are fine-tuned on ImageNet and evaluated on its 4 variants.}
    \label{tab:domain_gen}
\end{table*}
\vspace{-8pt}
\paragraph{Training Details:} All vanilla prompt-tuning and their ProTeCt counterparts are trained under the same setting. The following configuration is used unless noted. All experiments use SGD optimizer and the learning rate is set to 0.02 with a cosine learning rate scheduler. By default, a pretrained ViT-B/16 CLIP model is used as initialization. 
For Cifar100 and SUN, we train both CoOp and MaPLe prompts for 200 epochs, using a batch size of 128 and 32, respectively. For ImageNet, CoOp is trained for 30 epochs with a batch size of 8, while MaPLe is trained for 10 epochs with a batch size of 2. Note that the setting is slightly different from the original paper due to our GPU availability.

\vspace{-3pt}
\subsection{TOS Classification Performance} \vspace{-5pt}
Table~\ref{tab:main_table} shows that vanilla CoOp and MaPLe have reasonable leaf accuracy for both 1-shot and 16-shot classification on Cifar100, SUN, and ImageNet. However, their very low HCA shows that their predictions are not consistent over the class hierarchy. As a result, their TOS classification performance (MTA) is much weaker than their leaf accuracy. For example, 16-shot classification with CoOp on ImageNet has a leaf accuracy of 71.23, but expected TOS accuracy of 46.98. This is explained by the very low HCA of 2.99. Similar observations hold for different few-shot configurations. In all cases, ProTeCt (results on rows with a checkmark) significantly improves HCA and MTA(25). For example, it boosts the HCA of 16-shot classification with CoOp on ImageNet by 34.75 (2.99 vs 37.74), leading to an increase of MTA(25) of 41.63 (46.98 to 88.61). 

\vspace{-1pt}
Note that, in all cases, MTA(25) after ProTeCt training is {\it higher} than leaf accuracy. This is expected for a well-calibrated classifier, since decisions at intermediate levels of the tree are coarser-grained than those at the leaves, which can require very fine class distinctions. These results show that ProTeCt robustifies the model for use in the TOS classification setting. The table also shows that ProTeCt maintains leaf accuracies comparable to those of the vanilla methods.
Furthermore, the MTA results when 25 and 100 treecuts are sampled (corresponding to $T=5$ and $T=20$), are compared on Cifar100. It can be seen that the performances are similar, showing that sampling 25 treecuts is sufficient to achieve good estimation.
Fig.~\ref{fig:rel_gain_all_dataset} compares the \textbf{relative} gains in HCA and leaf accuracy of training with ProTeCt, as compared to vanilla prompt tuning. These gains are shown for both CoOp and MaPLe, under several few shot configurations, on SUN dataset. In all cases, ProTeCt increases HCA by more than 15 points, while maintaining a leaf accuracy comparable to that of vanilla CoOp/MaPLe. Similar results for Cifar100 and ImageNet can be found in appendix.

\vspace{-3pt}
\subsection{Domain Generalization of TOS Classification} \vspace{-3pt}
We investigate whether TOS classification performance generalizes across datasets, following the domain generalization setting of~\cite{coop,cocoop,maple,upt}. The CLIP model with ProTeCt prompts trained on ImageNet (source) is applied to 4 ImageNet variants (target) with visual domain shift: ImageNetv2~\cite{imagenetv2}, ImageNet-Sketch~\cite{imagenet_sketch}, ImageNet-A~\cite{imagenet_a} and ImageNet-R~\cite{imagenet_r}. Table~\ref{tab:domain_gen} summarizes the three metrics on these datasets for CoOp and MaPLe. Similarly to Table~\ref{tab:main_table}, ProTeCt enables significant gains in HCA and MTA(25) over the baselines for all datasets. Note that since ImageNet-A and ImageNet-R only contain 200 ImageNet subclasses, their hierarchy is different from that of ImageNet. These results demonstrate the flexibility and robustness of ProTeCt, even when transferring the model to a target domain whose class hierarchy is different from that of the source domain.

\vspace{-5pt}
\subsection{Ablation Study and Visualization} \vspace{-5pt}
In this section, we discuss the ablations of ProTeCt components and visualize the predictions (more in the appendix).

\vspace{-10pt}
\paragraph{\bf{Tree Dropout Rate $\beta$:}}
Fig.~\ref{fig:ablation} (a) plots Cifar100 $Acc_{leaf}$ and HCA as a function of the drop-out rate $\beta$, for 16-shot CoOp+ProTeCt training ($\lambda=1$). Larger values of $\beta$ reduce the likelihood of sampling the leaf nodes of the tree, resulting in shorter trees and weaker regularization. Hence, both leaf accuracy and HCA degrade for large $\beta$. However, always using the full tree ($\beta = 0$) also achieves sub-optimal results. The two metrics peak at $\beta=0.1$ and $\beta=0.2$, respectively. $\beta=0.1$ is selected for all experiments.

\begin{table}[t]
    \centering
    \resizebox{\linewidth}{!}{
    \begin{tabular}{|cc|ccc|ccc|}
      \hline
      \multirow{3}{*}{DTL} & \multirow{3}{*}{NCL} & \multicolumn{3}{c}{16-shot} & \multicolumn{3}{|c|}{1-shot} \\
        &  & $Acc_{Leaf}$ & HCA & MTA (25) & $Acc_{Leaf}$ & HCA & MTA (25) \\
      \hline
      \hline
       &  & 72.88 & 10.04 & 50.64 & 65.03 & 7.81 & 41.78 \\
      \checkmark &  & 72.81 & 47.97 & 87.32 & 64.77 & 32.93 & 81.38\\
       & \checkmark & 64.20 & 51.69 & 79.44 & 61.22 & 38.02 & 62.16\\
      \checkmark & \checkmark & \textbf{72.94} & \textbf{56.85} & \textbf{87.69} & \textbf{66.88} & \textbf{41.01} & \textbf{81.64} \\
      \hline
    \end{tabular}
    }
    \vspace{-8pt}
    \caption{Loss ablation with CoOp on Cifar100 dataset. Both losses improve the hierarchical consistency.}
    \label{tab:coop_cifar_ablade}
\end{table}
\begin{table}
    \centering
    \setlength{\tabcolsep}{4pt}
    \adjustbox{max width=\linewidth}{
    \begin{tabular}{|cc|ccc|ccc|}
    \hline
    K- & w/ & \multicolumn{3}{|c|}{CLIP-Adapter~\cite{CLIPAdapter}} & \multicolumn{3}{|c|}{CLIP+LORA~\cite{Doveh2022TeachingSV}} \\
    Shot & ProTeCt &  $Acc_{leaf}$ & HCA & MTA (25) & $Acc_{leaf}$ & HCA & MTA (25) \\
    \hline
    \hline
     16 & & 71.96 & 5.59 & 42.93 & 70.45 & 4.57 & 47.19\\
     16  & \checkmark &  72.47 & 57.15 & 87.67 & 70.64 & 51.06 & 77.29\\
     & &  {\color{red}(+0.51)} &  {\color{red}(+51.56)} &  {\color{red}(+44.83)} &  {\color{red}(+0.19)} &  {\color{red}(+46.49)} &  {\color{red}(+30.10)}\\
    \hline
     1 & & 65.35 & 8.35 & 48.25 & 63.57 & 2.89 & 38.63\\
     1  & \checkmark &   67.29 & 36.21 & 78.49 & 63.62 & 24.66 & 56.42\\
     & &  {\color{red}(+1.94)} &  {\color{red}(+27.86)} & {\color{red}(+30.24)} & {\color{red}(+0.05)} & {\color{red}(+21.8)} &  {\color{red}(+17.79)}\\
    \hline
    \end{tabular}
    }
    \vspace{-8pt}
    \caption{ProTeCt also improves adapter-based methods, including CLIP-Adapter~\cite{CLIPAdapter} and CLIP+LORA~\cite{Doveh2022TeachingSV} (dataset: Cifar100).}
    \label{tab:tuning_gen}
\end{table}

\vspace{-12pt}
\paragraph{\bf{Loss:}}
Fig.~\ref{fig:ablation}(b) ablates the strength of NCL loss (i.e. $\lambda$) for ProTeCt+CoOp using 1-shot setting on Cifar100 and $\beta=0.1$.
The introduction of NCL improves leaf accuracy/HCA from 64.8/32.9 ($\lambda=0$) to 66.9/41 ($\lambda=0.5$). We adopt $\lambda=0.5$ for CIFAR100 and SUN. For ImageNet, $\lambda=0.5$ and $\lambda=1$ have similar performance.
Table~\ref{tab:coop_cifar_ablade} further summarizes the CoOp+ProTeCt performance with and without the two losses of (\ref{eq:overallL}). Both losses improve TOS performance individually and there is a large additional gain when they are combined. Using NCL alone can degrade leaf performance, due to the lack of regularization across different levels of the hierarchy. The combination of the two losses overcomes this problem.

\vspace{-12pt}
\paragraph{\bf{Architecture:}}
Fig.~\ref{fig:ablation} (c) shows that the gains for CoOp+ProTeCt in Fig.~\ref{fig:rel_gain_all_dataset} with CLIP ViT B16 also hold for ViT B32, showing the plug-and-play properties of ProTeCt.

\vspace{-12pt}
\paragraph{\bf{Adapter-based tuning methods:}}
We further use the ProTeCt losses to train the CLIP adapter of~\cite{CLIPAdapter} and the CLIP+LORA method of~\cite{Doveh2022TeachingSV} to test the generation of ProTeCt.
Table~\ref{tab:tuning_gen} shows that this again produces large consistency gains on the TOS setting, indicating that ProTeCt losses generalize to both prompt-based and adapter-based methods.

\vspace{-12pt}
\paragraph{\bf{Visualization:}}
Fig.~\ref{fig:vis} shows examples from ImageNet (a,b) and its variants (c,d). While ProTeCt correctly classifies these examples at all hierarchy levels, vanilla prompt tuning fails at certain levels. More examples are in the appendix.

\vspace{-7pt}
\section{Conclusion} \vspace{-5pt}
In this work, we formulated the TOS classification setting, including datasets, performance metrics, and experiments. Given a dataset, a class hierarchy is built by assigning dataset classes to leaf nodes and superclasses to internal nodes. The TOS classifier is then expected to support classification with label sets drawn throughout the taxonomy. We have shown that existing FMs and prompting methods fail under this setting and proposed  ProTeCt training to enhance the TOS performance of FMs, as a plug-and-play method. ProTeCt includes two losses. A dynamic treecut loss, based on an efficient treecut sampler, dynamically regularizes labels of varying granularity. A node-centric loss encourages correct predictions at all hierarchy levels. Experiments show that ProTeCt enhances TOS performance of existing prompt-tuning techniques, and the gain generalizes across unseen domains. Finally, we show that ProTeCt is applicable to various architectures, hierarchies, and parameter-tuning methods.

\noindent\textbf{Acknowledgement}
This work was partially funded by NSF awards IIS-2303153, and a gift from Qualcomm. We also acknowledge and thank the use of the Nautilus platform for some of the experiments discussed above.

\clearpage
{
\small
\bibliographystyle{ieeenat_fullname}

}

\clearpage
\maketitlesupplementary

The appendix is organized as follows. Section~\ref{sec:train_details} provides more training details for ProTeCt. Section~\ref{sec:lemma_proof} shows the complete proof of Lemma 4.1. Section~\ref{sec:treecut_explain} shows more examples for explaining the implementation of treecut sampler. Section~\ref{sec:full_cifar}, Section~\ref{sec:full_sun}, and Section~\ref{sec:full_imagenet} shows the complete results conducted on Cifar100, Sun and ImageNet, respectively. Section~\ref{sec:full_imagenet} further shows the complete domain generalization results by applying the model trained on ImageNet to its 4 variants in a zero-shot fashion. We also test the robustness of ProTeCt on additional hierarchies in Section~\ref{sec:addtional}
with the FGVC Aircraft~\cite{Maji2013FineGrainedVC} dataset and the RSI-CB~\cite{li2020RSI-CB} satellite dataset.
Ablations of different ProTeCt components are shown in Section~\ref{sec:full_ablade} and more visualizations of incorrect predictions from existing prompt tuning methods are illustrated in Section~\ref{sec:vis}.

\section{Additional Training Details} \label{sec:train_details}
In addition to the training details provided in the main paper,  we list the url links that are used for training and evaluating ProTeCt. For CoOp
and CoCoOp
baselines, we adopt the code from \url{https://github.com/KaiyangZhou/CoOp}. For MaPLe, we adopt the code from \url{https://github.com/muzairkhattak/multimodal-prompt-learning}.

\section{Treecut size of a balanced M-ary tree } \label{sec:lemma_proof}
\begin{lemma2}
For a balanced M-ary tree with depth $L$ (root node is excluded and is at depth 0), the number of all valid treecut is $L + \sum_{l=2}^L \sum_{k=1}^{N-1} \frac{N!}{k!(N-k)!} |_{N=M^{l-1}}$
\end{lemma2}
\begin{proof}
This can be proved by induction. Given an M-ary tree with depth $L$, the number of treecuts is denoted as $f_L$. The idea is that when adding the depth $L$, we only need to recompute the additional possible treecuts between depth $L-1$ and $L$. Since there are $N=M^{L-1}$ nodes in layer $L-1$, the possible treecuts after adding layer $L$ is $1 + \sum_{k=1}^{N-1}\frac{N!}{k!(N-k)!} |_{N=M^{L-1}}$, where 1 indicates the treecut that covers all nodes at depth $L$ and $\sum_{k=1}^{N-1}\frac{N!}{k!(N-k)!} |_{N=M^{L-1}}$ means k nodes are covered in layer $L-1$. Below is the proof.

\begin{itemize}
\item When $L=1$, $f_1 = 1$.
\item When $L=2$, $f_2 = 1 + f_1 + \sum_{k=1}^{N-1}\frac{N!}{k!(N-k)!} |_{N=M^{L-1}}$. Consider the binary case, where $M=2$ and $N=M^{L-1}=2$, then $f_2= 1 + f_1 + \frac{2!}{1!(2-1)!} = 1 + 1 + 2 = 4$
\item Similarly, $f_3 = 1 + f_2 + \sum_{k=1}^{N-1}\frac{N!}{k!(N-k)!} |_{N=M^{L-1}}$
\item \begin{align}
    f_L &= 1 + f_{L-1}+\sum_{k=1}^{N-1}\frac{N!}{k!(N-k)!} |_{N=M^{L-1}} \nonumber\\
    &= 1 + 1 + f_{L-2}+\sum_{k=1}^{N-1}\frac{N!}{k!(N-k)!} |_{N=M^{L-2}} \nonumber\\
    &+\sum_{k=1}^{N-1}\frac{N!}{k!(N-k)!} |_{N=M^{L-1}}
    \nonumber\\
    &= L + \sum_{l=2}^L \sum_{k=1}^{N-1} \frac{N!}{k!(N-k)!} |_{N=M^{l-1}}
    \nonumber
\end{align}
\end{itemize}
\end{proof}

\section{Additional Examples for Treecut Sampler} \label{sec:treecut_explain}
\begin{figure*}[t!]
    \centering
    \includegraphics[width=0.85\linewidth]{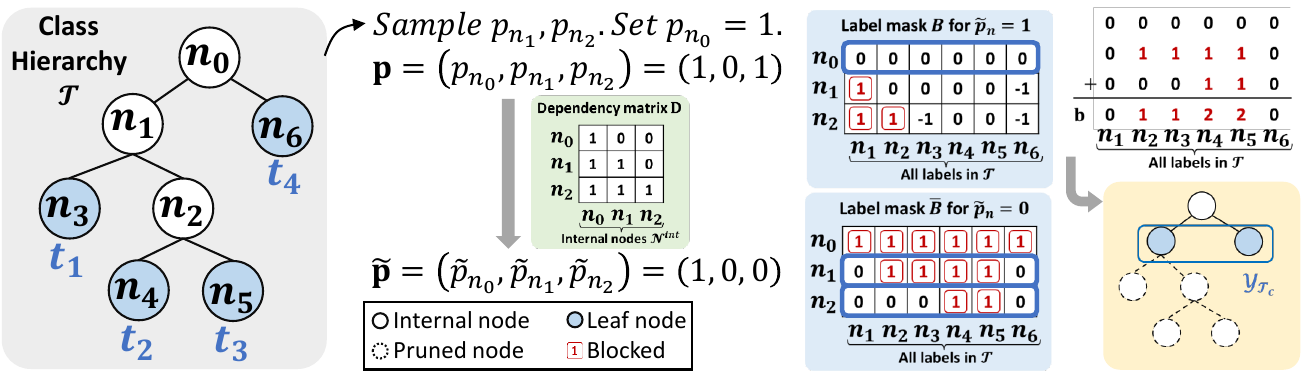}
    \captionof{figure}{Treecut example of $\mathcal{Y}_{\mathcal{T}_c}=\{n_1, n_6\}$.}\label{fig:ex1}
    \vspace{3pt}
    \includegraphics[width=0.85\linewidth]{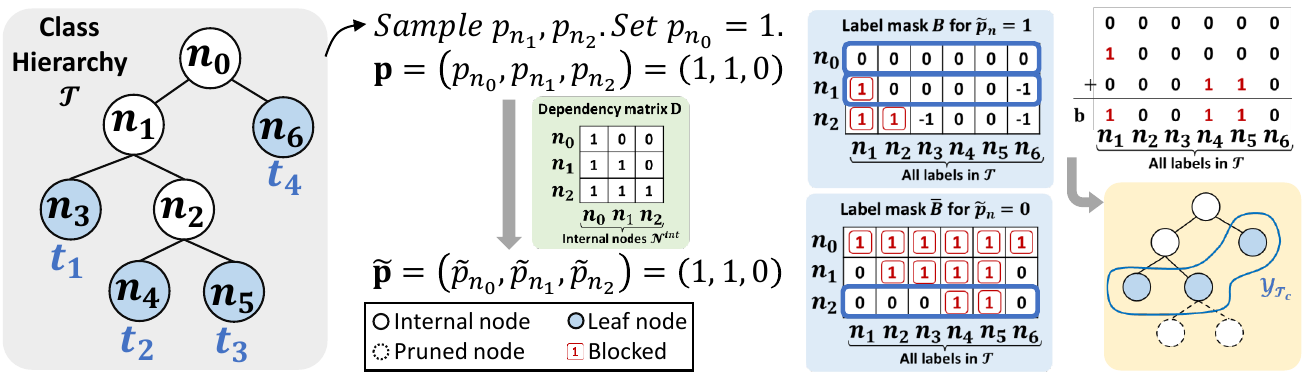}
    \captionof{figure}{Treecut example of $\mathcal{Y}_{\mathcal{T}_c}=\{n_2, n_3, n_6\}$.}\label{fig:ex2}
    \vspace{3pt}
    \includegraphics[width=0.85\linewidth]{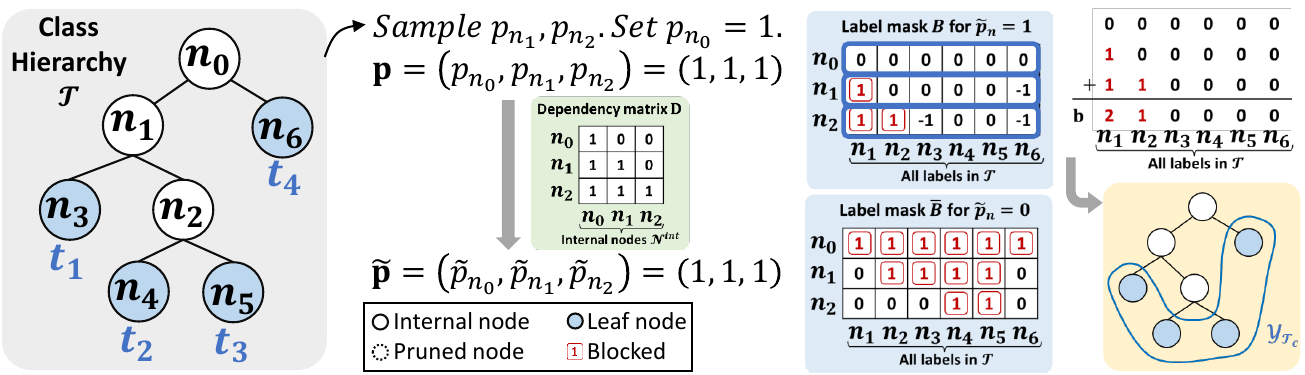}
    \captionof{figure}{Treecut example of $\mathcal{Y}_{\mathcal{T}_c}=\{n_3, n_4, n_5, n_6\}$.}\label{fig:ex3}
    \vspace{5pt}
\end{figure*}

In this section, we provide more detailed examples of the proposed Treecut sampler (i.e. Algorithm 1 in the main paper). Given the class hierarchy $\mathcal{T}$ on the left of Figure~\ref{fig:ex1}, 
three possible treecuts can be sampled 
by $\mathcal{T}$, i.e., $\mathcal{Y}_{\mathcal{T}_c}=\{n_1, n_6\}$ (see Figure~\ref{fig:ex1}), $\mathcal{Y}_{\mathcal{T}_c}=\{n_2, n_3, n_6\}$ (see Figure~\ref{fig:ex2}), and $\mathcal{Y}_{\mathcal{T}_c}=\{n_3, n_4, n_5, n_6\}$ (see Figure~\ref{fig:ex3}), depending on the sampled values $p_n$ at each internal node $n\in\mathcal{N}^{int}=\{n_0, n_1, n_2\}$. Note that $p_{n_0}$ is always set to 1 to ensure that the tree is not entirely pruned.
As described in the paper, we use a dependency matrix $\bf D$ to 
correct $\bf p$ as $\tilde{\bf p}$,
which is aligned with the dependency relationship among the internal nodes. For example, in the example shown in Figure~\ref{fig:ex1}, $p_{n_2}=1$ is corrected as $\tilde{p}_{n_2}=0$, since $n_2$ depends on $n_1$ and $p_{n_1}=0$.
A mask $\bf b$, flagging the unavailable labels, is then computed according to the values of $\tilde{\bf p}$. More specifically, the corresponding row in $min({\bf B}, 0)$ is fetched when $\tilde{p}_n=1$, and that in $\bar{\bf B}$ is used when $\tilde{p}_n=0$, for each internal node $n\in\mathcal{N}^{int}$. These masks are accumulated into the final mask $\bf b$, as shown on the right of each figure, where entries of 0 indicate the available labels for the sampled label set. For example, in Figure~\ref{fig:ex2}, the sampled label set contains $\{n_2, n_3, n_6\}$, because $b_2$, $b_3$ and $b_6$ are 0s.
Note that since the proposed Treecut Sampler maintains the node dependency with pre-computed matrices defined by the given hierarchy ${\cal T}$, it does not require any recursive traversal over the tree, and thus it is very efficient for the {\it on-the-fly} treecut sampling.

\section{Complete Table of Cifar100 Experiments} \label{sec:full_cifar}
In this section, we report the complete experiment results conducted on Cifar100.
Table~\ref{tab:full_cifar100_coop_vitb16}, Table~\ref{tab:full_cifar100_coop_vitb32} and Table~\ref{tab:full_cifar100_coop_vitl14} shows the results of vanilla CoOp and its results after adding ProTeCt. The CLIP features from ViT B16, ViT B32 and ViT L14 are considered in Table~\ref{tab:full_cifar100_coop_vitb16}, Table~\ref{tab:full_cifar100_coop_vitb32} and Table~\ref{tab:full_cifar100_coop_vitl14}, respectively. While it is known that CLIP ViT L14 has a more powerful representation than ViT B32 and ViT B16 (also reflected in the leaf accuracy between three tables), all of them perform equally poor in terms of HCA (10.04/4.95/11.14 for 16-shot CoOp using CLIP B16/B32/L14 feature). This shows that simply using a stronger CLIP feature does not address the problem of hierarchical classification and does not improve hierarchical consistency. Furthermore, Table~\ref{tab:full_cifar100_coop_vitb16} contains the result of ProTeCt without using the treecut sampler ($\beta=0$ ; Block 2 and Block 3) and without using NCL loss of (7) ($\lambda=0$ ; Block 4) under multiple low-shot settings. For example, when 16-shot is considered, adding both NCL loss and treecut sampler ($\lambda=0.5$ and $\beta=0.1$) gives the result of 56.85 for HCA. Removing the tree dropout ($\lambda=0.5$ and $\beta=0$) yields 51.99 and removing the NCL loss ($\lambda=0$ and $\beta=0.1$) yields 47.97. This shows that both the NCL loss and the treecut sampler are important and lead to a significant gain over vanilla CoOp (HCA=10.04). Table~\ref{tab:full_cifar100_maple} shows similar results when adding ProTeCt on MaPLe. Furthermore, we sampled $T$ treecuts for each dropout rate $\beta = \{0.1, 0.3, 0.5, 0.7, 0.9\}$, where $T=5$ and $T=20$, resulting in 25 and 100 treecuts, respectively. 
Table~\ref{tab:full_cifar100_treecut} demonstrates ProTeCt can improve the MTA metric for both CoOp and MaPLe for both 25 and 100 randomly sampled treecuts.
Table~\ref{tab:resnet} further shows that ProTeCt can generalize to ResNet-based architectures.

\begin{table}[t!]
    \centering
    \adjustbox{max width=0.9\linewidth}{
    \begin{tabular}{|cccccc|cc|}
    \hline
    Method  & Encoder & K-shot & w/ ProTeCt & $\lambda$ & $\beta$  & Leaf Acc. & HCA \\
    \hline
    \hline
    CoOp & ViT B16 & 16 &  & N/A & N/A &  72.88 $\pm$ 0.62 & 10.04 $\pm$ 1.11 \\
    CoOp & ViT B16 & 8  &  & N/A & N/A &  70.84	$\pm$ 0.85 & 6.03 $\pm$ 0.64\\
    CoOp & ViT B16 & 4  &  & N/A & N/A &  69.47	$\pm$ 0.90 & 6.15 $\pm$ 1.04\\
    CoOp & ViT B16 & 2  &  & N/A & N/A &  68.17	$\pm$ 0.57 & 4.19 $\pm$ 0.81\\
    CoOp & ViT B16 & 1  &  & N/A & N/A &  65.03	$\pm$ 0.56 & 7.81 $\pm$ 0.14\\
    \hline
    CoOp & ViT B16 & 16 & \checkmark & 0.5 & 0 &  72.08 $\pm$ 0.38	& 51.99 $\pm$ 0.24  \\
    CoOp & ViT B16 & 8  & \checkmark & 0.5 & 0 &  68.94	$\pm$ 0.52	& 49.01	$\pm$ 0.54 \\
    CoOp & ViT B16 & 4  & \checkmark & 0.5 & 0 &  66.38	$\pm$ 1.18	& 45.24	$\pm$ 0.93 \\
    CoOp & ViT B16 & 2  & \checkmark & 0.5 & 0 &  63.96	$\pm$ 0.57	& 42.78	$\pm$ 1.49 \\
    CoOp & ViT B16 & 1  & \checkmark & 0.5 & 0 &  62.01	$\pm$ 0.80	& 34.90	$\pm$ 1.08 \\
    \hline
    CoOp & ViT B16 & 16 & \checkmark & 1 & 0 &  70.86 $\pm$ 0.59 &	54.39 $\pm$ 0.68  \\
    CoOp & ViT B16 & 8  & \checkmark & 1 & 0 &  68.76 $\pm$ 0.90 &	52.14 $\pm$ 0.32 \\
    CoOp & ViT B16 & 4  & \checkmark & 1 & 0 &  66.92 $\pm$ 0.20 &	47.63 $\pm$ 0.54 \\
    CoOp & ViT B16 & 2  & \checkmark & 1 & 0 &  64.87 $\pm$ 1.28 &	40.74 $\pm$ 0.87 \\
    CoOp & ViT B16 & 1  & \checkmark & 1 & 0 &  62.57 $\pm$ 0.06 &	38.97 $\pm$ 1.29 \\
    \hline
    CoOp & ViT B16 & 16 & \checkmark & 0 & 0.1 & 72.81 $\pm$ 0.31 & 47.97 $\pm$ 0.70   \\
    CoOp & ViT B16 & 8  & \checkmark & 0 & 0.1 & 70.94 $\pm$ 0.18 & 48.53 $\pm$ 0.02 \\
    CoOp & ViT B16 & 4  & \checkmark & 0 & 0.1 & 69.10 $\pm$ 0.92 & 45.20 $\pm$ 0.25 \\
    CoOp & ViT B16 & 2  & \checkmark & 0 & 0.1 & 68.85 $\pm$ 0.11 & 42.28 $\pm$ 1.57 \\
    CoOp & ViT B16 & 1  & \checkmark & 0 & 0.1 & 64.77 $\pm$ 1.37 & 32.93 $\pm$ 0.42 \\
    \hline
    CoOp & ViT B16 & 16  & \checkmark & 0.5 & 0.1 & 72.94 $\pm$ 0.83 & 56.85 $\pm$ 1.60 \\
    CoOp & ViT B16 & 8  & \checkmark & 0.5 & 0.1 & 71.10 $\pm$ 1.06 & 52.27 $\pm$ 0.62 \\
    CoOp & ViT B16 & 4  & \checkmark & 0.5 & 0.1 & 69.46 $\pm$ 0.58 & 48.71 $\pm$ 0.13 \\
    CoOp & ViT B16 & 2 & \checkmark & 0.5 & 0.1 & 68.63 $\pm$ 0.67 & 46.03 $\pm$ 0.24  \\
    CoOp & ViT B16 & 1  & \checkmark & 0.5 & 0.1 & 66.88 $\pm$ 0.21 & 41.01 $\pm$ 1.18 \\
    \hline
    CoOp & ViT B16 & 16 & \checkmark & 1 & 0.1 & 73.26 $\pm$ 0.66 & 58.01 $\pm$ 0.43  \\
    CoOp & ViT B16 & 8  & \checkmark & 1 & 0.1 & 70.10 $\pm$ 0.08 & 52.81 $\pm$ 0.05 \\
    CoOp & ViT B16 & 4  & \checkmark & 1 & 0.1 & 68.41 $\pm$ 0.50 & 49.59 $\pm$ 0.89 \\
    CoOp & ViT B16 & 2  & \checkmark & 1 & 0.1 & 67.73 $\pm$ 1.25 & 45.27 $\pm$ 0.28 \\
    CoOp & ViT B16 & 1  & \checkmark & 1 & 0.1 & 63.84 $\pm$ 1.51 & 40.05 $\pm$ 1.48 \\
    \hline
    \end{tabular}
    }
    \caption{Performance of few-shot CoOp on Cifar100 under  ViT B16. Ablations cover both NCL strengths $\lambda$ and tree dropout rate $\beta$.}
    \label{tab:full_cifar100_coop_vitb16}
\end{table}

\begin{table}[t!]
    \centering
    \adjustbox{max width=0.9\linewidth}{
    \begin{tabular}{|cccccc|cc|}
    \hline
    Method  & Encoder & K-shot & w/ ProTeCt & $\lambda$ & $\beta$  & Leaf Acc. & HCA \\
    \hline
    \hline
    CoOp & ViT B32 & 16 &  & N/A & N/A &  68.13 $\pm$ 0.19 & 4.95 $\pm$  0.61 \\
    CoOp & ViT B32 & 8  &  & N/A & N/A &  65.52	$\pm$ 0.15 & 5.82 $\pm$  0.29\\
    CoOp & ViT B32 & 4  &  & N/A & N/A &  63.42 $\pm$ 1.40 & 8.56 $\pm$  0.72\\
    CoOp & ViT B32 & 2  &  & N/A & N/A &  63.65	$\pm$ 0.60 & 10.25 $\pm$ 0.88\\
    CoOp & ViT B32 & 1  &  & N/A & N/A &  59.53	$\pm$ 0.60 & 3.43 $\pm$  0.86\\
    \hline
    CoOp & ViT B32 & 16 & \checkmark & 0 & 0.1 & 68.42 $\pm$ 0.91 & 47.79 $\pm$ 0.54   \\
    CoOp & ViT B32 & 8  & \checkmark & 0 & 0.1 & 66.39 $\pm$ 0.48 & 44.47 $\pm$ 0.98 \\
    CoOp & ViT B32 & 4  & \checkmark & 0 & 0.1 & 64.73 $\pm$ 0.17 & 31.72 $\pm$ 0.33 \\
    CoOp & ViT B32 & 2  & \checkmark & 0 & 0.1 & 64.55 $\pm$ 0.44 & 30.78 $\pm$ 0.66 \\
    CoOp & ViT B32 & 1  & \checkmark & 0 & 0.1 & 60.91 $\pm$ 0.42 & 34.64 $\pm$ 0.55 \\
    \hline
    CoOp & ViT B32 & 16  & \checkmark & 0.5 & 0.1 & 68.87 $\pm$ 1.09 & 51.55 $\pm$ 0.65 \\
    CoOp & ViT B32 & 8  & \checkmark & 0.5 & 0.1 & 66.85 $\pm$ 0.32 & 48.39 $\pm$ 1.35 \\
    CoOp & ViT B32 & 4  & \checkmark & 0.5 & 0.1 & 65.41 $\pm$ 0.74 & 41.63 $\pm$ 0.39 \\
    CoOp & ViT B32 & 2 & \checkmark & 0.5 & 0.1 & 62.86 $\pm$ 0.81 & 38.13 $\pm$ 0.61  \\
    CoOp & ViT B32 & 1  & \checkmark & 0.5 & 0.1 & 61.59 $\pm$ 0.80 & 35.65 $\pm$ 0.19 \\
    \hline
    CoOp & ViT B32 & 16 & \checkmark & 1 & 0.1 & 68.93 $\pm$ 0.22 & 51.67 $\pm$ 0.58  \\
    CoOp & ViT B32 & 8  & \checkmark & 1 & 0.1 & 65.54 $\pm$ 0.54 & 48.36 $\pm$ 0.63 \\
    CoOp & ViT B32 & 4  & \checkmark & 1 & 0.1 & 64.28 $\pm$ 0.07 & 42.78 $\pm$ 1.04 \\
    CoOp & ViT B32 & 2  & \checkmark & 1 & 0.1 & 61.68 $\pm$ 0.67 & 40.53 $\pm$ 0.42 \\
    CoOp & ViT B32 & 1  & \checkmark & 1 & 0.1 & 58.98 $\pm$ 0.88 & 36.59 $\pm$ 0.76 \\
    \hline
    \end{tabular}
    }
    \caption{Performance of few-shot CoOp on Cifar100 under ViT B32. Ablations cover different NCL strengths $\lambda$.}
    \label{tab:full_cifar100_coop_vitb32}
\end{table}

\begin{table}[t!]
    \centering
    \adjustbox{max width=0.9\linewidth}{
    \begin{tabular}{|cccccc|cc|}
    \hline
    Method  & Encoder & K-shot & w/ ProTeCt & $\lambda$ & $\beta$  & Leaf Acc. & HCA \\
    \hline
    \hline
    CoOp & ViT L14 & 16 &  & N/A & N/A &  79.98 $\pm$ 0.97 & 11.14 $\pm$  0.47 \\
    CoOp & ViT L14 & 8  &  & N/A & N/A &  79.37	$\pm$ 0.90 & 6.91 $\pm$  0.67\\
    CoOp & ViT L14 & 4  &  & N/A & N/A &  77.34 $\pm$ 0.78 & 7.78 $\pm$  0.82\\
    CoOp & ViT L14 & 2  &  & N/A & N/A &  76.63	$\pm$ 0.65 & 5.21  $\pm$ 0.87\\
    CoOp & ViT L14 & 1  &  & N/A & N/A &  73.26	$\pm$ 0.95 & 4.87 $\pm$  0.15\\
    \hline
    CoOp & ViT L14 & 16 & \checkmark & 0 & 0.1 & 81.17 $\pm$ 0.34 & 63.40 $\pm$ 0.30   \\
    CoOp & ViT L14 & 8  & \checkmark & 0 & 0.1 & 80.00  $\pm$ 0.98 & 62.11 $\pm$ 0.81 \\
    CoOp & ViT L14 & 4  & \checkmark & 0 & 0.1 & 79.05 $\pm$ 0.68 & 57.19 $\pm$ 0.26 \\
    CoOp & ViT L14 & 2  & \checkmark & 0 & 0.1 & 78.53 $\pm$ 0.69 & 40.59 $\pm$ 0.68 \\
    CoOp & ViT L14 & 1  & \checkmark & 0 & 0.1 & 76.48 $\pm$ 0.52 & 45.11 $\pm$ 0.68 \\
    \hline
    CoOp & ViT L14 & 16  & \checkmark & 0.5 & 0.1 & 80.95 $\pm$ 0.38 & 68.92 $\pm$ 0.77 \\
    CoOp & ViT L14 & 8  & \checkmark & 0.5 & 0.1 & 79.87 $\pm$ 0.11 & 64.05 $\pm$ 0.57 \\
    CoOp & ViT L14 & 4  & \checkmark & 0.5 & 0.1 & 79.18 $\pm$ 0.51 & 51.88 $\pm$ 0.45 \\
    CoOp & ViT L14 & 2 & \checkmark & 0.5 & 0.1 & 76.76 $\pm$ 0.24 & 51.96 $\pm$ 0.06  \\
    CoOp & ViT L14 & 1  & \checkmark & 0.5 & 0.1 & 73.89 $\pm$ 0.62 & 50.31 $\pm$ 1.02 \\
    \hline
    CoOp & ViT L14 & 16 & \checkmark & 1 & 0.1 & 80.45 $\pm$ 0.90 & 70.15 $\pm$ 0.98  \\
    CoOp & ViT L14 & 8  & \checkmark & 1 & 0.1 & 79.25 $\pm$ 0.93 & 65.75 $\pm$ 0.69 \\
    CoOp & ViT L14 & 4  & \checkmark & 1 & 0.1 & 78.37 $\pm$ 0.13 & 47.30 $\pm$ 0.20 \\
    CoOp & ViT L14 & 2  & \checkmark & 1 & 0.1 & 75.21 $\pm$ 0.33 & 54.78 $\pm$ 0.66 \\
    CoOp & ViT L14 & 1  & \checkmark & 1 & 0.1 & 74.93 $\pm$ 0.15 & 52.08 $\pm$ 1.04 \\
    \hline
    \end{tabular}
    }
    \caption{Performance of few-shot CoOp on Cifar100 under both ViT L14. Ablations cover different NCL strengths $\lambda$.}
    \label{tab:full_cifar100_coop_vitl14}
\end{table}

\begin{table}[t!]
    \centering
    \adjustbox{max width=0.9\linewidth}{
    \begin{tabular}{|cccccc|cc|}
    \hline
    Method  & Encoder & K-shot & w/ ProTeCt & $\lambda$ & $\beta$  & Leaf Acc. & HCA \\
    \hline
    \hline
    MaPLe & ViT B16 & 16 &  & N/A & N/A & 75.01 $\pm$ 0.37 & 17.54 $\pm$ 0.83  \\
    MaPLe & ViT B16 & 8  &  & N/A & N/A & 73.93 $\pm$ 0.46 & 9.44 $\pm$ 1.13  \\
    MaPLe & ViT B16 & 4  &  & N/A & N/A & 72.68 $\pm$ 0.47 & 20.29 $\pm$ 1.07 \\
    MaPLe & ViT B16 & 2  &  & N/A & N/A & 71.37 $\pm$ 1.39 & 12.15 $\pm$ 0.25 \\
    MaPLe & ViT B16 & 1  &  & N/A & N/A & 68.75 $\pm$ 0.96 & 4.65 $\pm$ 1.52 \\
    \hline
    MaPLe & ViT B16 & 16 & \checkmark & 0 & 0.1 &  75.82 $\pm$ 0.10 & 58.63 $\pm$ 0.43  \\
    MaPLe & ViT B16 & 8  & \checkmark & 0 & 0.1 &  74.29 $\pm$ 0.91 & 57.31 $\pm$ 0.79  \\
    MaPLe & ViT B16 & 4  & \checkmark & 0 & 0.1 &  72.92 $\pm$ 0.42 & 54.12 $\pm$ 1.56  \\
    MaPLe & ViT B16 & 2  & \checkmark & 0 & 0.1 &  71.09 $\pm$ 1.35 & 47.78 $\pm$ 0.64  \\
    MaPLe & ViT B16 & 1  & \checkmark & 0 & 0.1 &  68.32 $\pm$ 0.20 & 39.43 $\pm$ 0.25 \\
    \hline
    MaPLe & ViT B16 & 16 & \checkmark & 0.5 & 0.1 & 75.34 $\pm$ 0.39 & 61.15 $\pm$ 0.53\\
    MaPLe & ViT B16 & 8  & \checkmark & 0.5 & 0.1 & 74.30 $\pm$ 0.29 & 60.24 $\pm$ 0.82\\
    MaPLe & ViT B16 & 4  & \checkmark & 0.5 & 0.1 & 71.35 $\pm$ 0.61 & 56.03 $\pm$ 0.35\\
    MaPLe & ViT B16 & 2  & \checkmark & 0.5 & 0.1 & 70.24 $\pm$ 1.01 & 52.56 $\pm$ 0.48 \\
    MaPLe & ViT B16 & 1  & \checkmark & 0.5 & 0.1 & 69.33 $\pm$ 0.81 & 48.10 $\pm$ 0.26 \\
    \hline
    MaPLe & ViT B16 & 16 & \checkmark & 1 & 0.1 & 76.30 $\pm$ 0.56 & 62.04 $\pm$ 0.97  \\
    MaPLe & ViT B16 & 8  & \checkmark & 1 & 0.1 & 73.60 $\pm$ 0.69 & 61.20 $\pm$ 0.77 \\
    MaPLe & ViT B16 & 4  & \checkmark & 1 & 0.1 & 72.06 $\pm$ 0.34 & 56.51 $\pm$ 1.24 \\
    MaPLe & ViT B16 & 2  & \checkmark & 1 & 0.1 & 69.95 $\pm$ 1.30 & 53.53 $\pm$ 0.67 \\
    MaPLe & ViT B16 & 1  & \checkmark & 1 & 0.1 & 70.44 $\pm$ 0.10 & 46.94 $\pm$ 0.85 \\
    \hline
    \end{tabular}
    }
    \caption{Performance of few-shot MaPLe on Cifar100. Ablations cover different NCL strengths $\lambda$.}
    \label{tab:full_cifar100_maple}
\end{table}

\begin{table}[t!]
    \centering
    \adjustbox{max width=0.9\linewidth}{
    \begin{tabular}{|cccccc|cc|}
    \hline
    Method  & Encoder & K-shot & w/ ProTeCt & $\lambda$ & $\beta$  & MTA (25) &  MTA (100)\\
    \hline
    \hline
    CoOp & ViT B32 & 16 &  & N/A & N/A & 52.33 & 54.58 \\
    CoOp & ViT B32 & 8  &  & N/A & N/A & 46.09 & 47.20 \\
    CoOp & ViT B32 & 4  &  & N/A & N/A & 53.35 & 54.30 \\
    CoOp & ViT B32 & 2  &  & N/A & N/A & 53.13 & 53.81 \\
    CoOp & ViT B32 & 1  &  & N/A & N/A & 38.80 & 40.16 \\
    \hline
    CoOp & ViT B32 & 16 & \checkmark & 0.5 & 0.1 & 86.26 & 85.73 \\
    CoOp & ViT B32 & 8  & \checkmark & 0.5 & 0.1 & 85.05 & 84.57 \\
    CoOp & ViT B32 & 4  & \checkmark & 0.5 & 0.1 & 81.01 & 80.61 \\
    CoOp & ViT B32 & 2  & \checkmark & 0.5 & 0.1 & 79.95 & 79.98 \\
    CoOp & ViT B32 & 1  & \checkmark & 0.5 & 0.1 & 78.08 & 76.95 \\
    \hline
    \hline
    CoOp & ViT B16 & 16 &  & N/A & N/A & 50.64 & 51.14 \\
    CoOp & ViT B16 & 8  &  & N/A & N/A & 47.95 & 50.41 \\
    CoOp & ViT B16 & 4  &  & N/A & N/A & 43.77 & 46.29 \\
    CoOp & ViT B16 & 2  &  & N/A & N/A & 40.81 & 42.95 \\
    CoOp & ViT B16 & 1  &  & N/A & N/A & 41.78 & 44.17 \\
    \hline
    CoOp & ViT B16 & 16 & \checkmark & 0.5 & 0.1 & 87.69 & 87.30 \\
    CoOp & ViT B16 & 8  & \checkmark & 0.5 & 0.1 & 86.28 & 86.01 \\
    CoOp & ViT B16 & 4  & \checkmark & 0.5 & 0.1 & 84.52 & 83.79 \\
    CoOp & ViT B16 & 2  & \checkmark & 0.5 & 0.1 & 83.49 & 83.18 \\
    CoOp & ViT B16 & 1  & \checkmark & 0.5 & 0.1 & 81.64 & 81.01 \\
    \hline
    \hline
    CoOp & ViT L14 & 16 &  & N/A & N/A & 58.81 & 60.89 \\
    CoOp & ViT L14 & 8  &  & N/A & N/A & 40.49 & 43.20 \\
    CoOp & ViT L14 & 4  &  & N/A & N/A & 44.71 & 47.39 \\
    CoOp & ViT L14 & 2  &  & N/A & N/A & 39.44 & 43.22 \\
    CoOp & ViT L14 & 1  &  & N/A & N/A & 52.32 & 54.90 \\
    \hline
    CoOp & ViT L14 & 16 & \checkmark & 0.5 & 0.1 & 90.83 & 90.48 \\
    CoOp & ViT L14 & 8  & \checkmark & 0.5 & 0.1 & 89.39 & 89.16 \\
    CoOp & ViT L14 & 4  & \checkmark & 0.5 & 0.1 & 84.48 & 84.79 \\
    CoOp & ViT L14 & 2  & \checkmark & 0.5 & 0.1 & 85.57 & 85.29 \\
    CoOp & ViT L14 & 1  & \checkmark & 0.5 & 0.1 & 83.65 & 83.52 \\
    \hline
    \hline
    MaPLe & ViT B16 & 16 &  & N/A & N/A & 52.21 & 50.82\\
    MaPLe & ViT B16 & 8  &  & N/A & N/A & 58.56 & 61.48\\
    MaPLe & ViT B16 & 4  &  & N/A & N/A & 66.14 & 67.06\\
    MaPLe & ViT B16 & 2  &  & N/A & N/A & 55.98 & 57.59\\
    MaPLe & ViT B16 & 1  &  & N/A & N/A & 50.60 & 54.99\\
    \hline
    MaPLe & ViT B16 & 16 & \checkmark & 0.5 & 0.1 & 88.04 & 88.33 \\
    MaPLe & ViT B16 & 8  & \checkmark & 0.5 & 0.1 & 87.65 & 88.13 \\
    MaPLe & ViT B16 & 4  & \checkmark & 0.5 & 0.1 & 86.72 & 87.04 \\
    MaPLe & ViT B16 & 2  & \checkmark & 0.5 & 0.1 & 85.03 & 85.39 \\
    MaPLe & ViT B16 & 1  & \checkmark & 0.5 & 0.1 & 83.36 & 83.78 \\
    \hline
    \end{tabular}
    }
    \caption{Performance of MTA for both few-shot CoOp and MaPLe on Cifar100. 25 ($T=5$) and 100 ($T=20$) treecuts are sampled for MTA evaluation.}
    \label{tab:full_cifar100_treecut}
\end{table}

\begin{table}[b]
    \centering
    \resizebox{0.9\linewidth}{!}{
    \begin{tabular}{|cccc|ccc|}
    \hline
    Method & Encoder & K-shot & w/ ProTeCt & $Acc_{leaf}$ & HCA & MTA (25) \\ \hline\hline
    CoOp & ResNet-50 & 16 & & 52.61 & 5.72 & 41.97 \\
    CoOp & ResNet-50 & 16 & \checkmark & 52.83 & 33.34 & 79.06 \\ \hline
    CoOp & ResNet-101 & 16 & & 56.97 & 5.58 & 53.43 \\
    CoOp & ResNet-101 & 16 & \checkmark & 57.64 & 39.93 & 81.76 \\
    \hline
    \end{tabular}
    }
    \caption{CoOp 16-shot results on Cifar100 with ResNets.}
    \label{tab:resnet}
\end{table}

\clearpage
\section{Complete Table of SUN Experiments}\label{sec:full_sun}
In this section, we report the complete experiment result conducted on SUN.
Table~\ref{tab:full_sun_coop} and Table~\ref{tab:full_sun_maple} show the results of vanilla CoOp and MaPLe, and their results after adding ProTeCt. When comparing the HCA results of the vanilla prompt tuning with that of Cifar100 and ImageNet, the HCA result on SUN is much higher and the gap between HCA and $Acc_{leaf}$ is much smaller. This is due to the shallow hierarchy of SUN dataset, indicating SUN is a much simpler dataset for hierarchical classification. However, we still see that ProTeCt achieves consistent improvement over the vanilla prompt tuning methods.
Table~\ref{tab:full_sun_treecut} further compares the MTA result of vanilla CoOp and MaPLe, and their ProTeCt counterpart.

\begin{table}[H]
    \centering
    \adjustbox{max width=0.9\linewidth}{
    \begin{tabular}{|cccccc|cc|}
    \hline
    Method  & Encoder & K-shot & w/ ProTeCt & $\lambda$ & $\beta$ & Leaf Acc. & HCA \\
    \hline
    \hline
    CoOp & ViT B16 & 16 &  & N/A & N/A &  73.82 $\pm$ 0.12	& 38.28 $\pm$ 0.46    \\
    CoOp & ViT B16 & 8  &  & N/A & N/A &  71.77 $\pm$ 0.67	& 33.95 $\pm$ 0.08     \\
    CoOp & ViT B16 & 4  &  & N/A & N/A &  69.31 $\pm$ 0.51	& 30.51 $\pm$ 0.71     \\
    CoOp & ViT B16 & 2  &  & N/A & N/A &  66.34 $\pm$ 0.33	& 36.85 $\pm$ 0.67     \\
    CoOp & ViT B16 & 1  &  & N/A & N/A &  63.65 $\pm$ 1.42	& 33.36 $\pm$ 0.21      \\
    \hline
    CoOp & ViT B16 & 16 & \checkmark & 0 & 0.1 & 74.95 $\pm$ 0.69 & 60.95 $\pm$ 0.91 \\
    CoOp & ViT B16 & 8  & \checkmark & 0 & 0.1 & 72.31 $\pm$ 0.18 & 57.61 $\pm$ 1.31 \\
    CoOp & ViT B16 & 4  & \checkmark & 0 & 0.1 & 69.53 $\pm$ 0.77 & 54.79 $\pm$ 0.12 \\
    CoOp & ViT B16 & 2  & \checkmark & 0 & 0.1 & 67.01 $\pm$ 1.10 &  50.78  $\pm$ 0.03\\
    CoOp & ViT B16 & 1  & \checkmark & 0 & 0.1 & 64.45 $\pm$ 0.96 & 47.75 $\pm$ 0.11 \\
    \hline
    CoOp & ViT B16 & 8  & \checkmark & 0.5 & 0.1 & 74.59 $\pm$ 0.41 & 62.94 $\pm$ 0.15\\
    CoOp & ViT B16 & 4  & \checkmark & 0.5 & 0.1 & 71.53 $\pm$ 0.67 & 58.17 $\pm$ 0.33\\
    CoOp & ViT B16 & 2  & \checkmark & 0.5 & 0.1 & 69.80 $\pm$ 0.98 & 56.85 $\pm$ 0.41\\
    CoOp & ViT B16 & 16 & \checkmark & 0.5 & 0.1 & 67.29 $\pm$ 1.32 & 51.82 $\pm$ 1.20 \\
    CoOp & ViT B16 & 1  & \checkmark & 0.5 & 0.1 & 63.79 $\pm$ 1.16 & 49.62 $\pm$ 1.40\\
    \hline
    CoOp & ViT B16 & 16 & \checkmark & 1 & 0.1 & 74.31 $\pm$ 0.23 & 62.96 $\pm$ 0.61 \\
    CoOp & ViT B16 & 8  & \checkmark & 1 & 0.1 & 71.27 $\pm$ 0.42 & 58.74 $\pm$ 0.98 \\
    CoOp & ViT B16 & 4  & \checkmark & 1 & 0.1 & 68.81 $\pm$ 0.71 & 55.90 $\pm$ 0.09 \\
    CoOp & ViT B16 & 2  & \checkmark & 1 & 0.1 & 67.66 $\pm$ 0.51 & 50.94 $\pm$ 1.31 \\
    CoOp & ViT B16 & 1  & \checkmark & 1 & 0.1 & 63.95 $\pm$ 1.19 & 50.99 $\pm$ 1.21 \\
    \hline
    \end{tabular}
    }
    \caption{Performance of few-shot CoOp on SUN. Ablations cover different NCL strengths $\lambda$.}
    \label{tab:full_sun_coop}
\end{table}

\begin{table}[H]
    \centering
\adjustbox{max width=0.9\linewidth}{
    \begin{tabular}{|cccccc|cc|}
    \hline
    Method  & Encoder & K-shot & w/ ProTeCt & $\lambda$ & $\beta$ & Leaf Acc. & HCA \\
    \hline
    \hline
    MaPLe & ViT B16 & 16 &  & N/A & N/A & 71.86	$\pm$ 0.11 & 33.25 $\pm$ 1.31  \\
    MaPLe & ViT B16 & 8  &  & N/A & N/A & 68.96	$\pm$ 0.51 & 29.63 $\pm$ 0.19\\
    MaPLe & ViT B16 & 4  &  & N/A & N/A & 67.27	$\pm$ 0.45 & 25.97 $\pm$ 0.53\\
    MaPLe & ViT B16 & 2  &  & N/A & N/A & 65.33	$\pm$ 1.21 & 29.79 $\pm$ 0.13\\
    MaPLe & ViT B16 & 1  &  & N/A & N/A & 63.98	$\pm$ 0.99 & 25.15 $\pm$ 0.76\\
    \hline
    MaPLe & ViT B16 & 16 & \checkmark & 0 & 0.1 &  72.89 $\pm$ 0.77 & 56.52 $\pm$ 0.88  \\
    MaPLe & ViT B16 & 8  & \checkmark & 0 & 0.1 &  71.24 $\pm$ 0.76 & 55.49 $\pm$ 1.05 \\
    MaPLe & ViT B16 & 4  & \checkmark & 0 & 0.1 &  69.24 $\pm$ 0.41 & 51.88 $\pm$ 1.22 \\
    MaPLe & ViT B16 & 2  & \checkmark & 0 & 0.1 &  66.98 $\pm$ 0.44 & 51.60 $\pm$ 0.55 \\
    MaPLe & ViT B16 & 1  & \checkmark & 0 & 0.1 &  63.80 $\pm$ 1.51 & 47.93 $\pm$ 0.31 \\
    \hline
    MaPLe & ViT B16 & 16 & \checkmark & 0.5 & 0.1 & 72.17 $\pm$ 1.20 &  59.71  $\pm$ 0.04 \\
    MaPLe & ViT B16 & 8  & \checkmark & 0.5 & 0.1 & 71.04 $\pm$ 0.09 & 57.78 $\pm$ 1.22 \\
    MaPLe & ViT B16 & 4  & \checkmark & 0.5 & 0.1 & 68.64 $\pm$ 0.61 & 54.86 $\pm$ 1.08 \\
    MaPLe & ViT B16 & 2  & \checkmark & 0.5 & 0.1 & 66.37 $\pm$ 0.62 & 53.13 $\pm$ 0.39 \\
    MaPLe & ViT B16 & 1  & \checkmark & 0.5 & 0.1 & 64.29 $\pm$ 1.23 & 50.45 $\pm$ 0.40 \\
    \hline
    MaPLe & ViT B16 & 16 & \checkmark & 1 & 0.1 & 71.03 $\pm$ 0.99 & 59.92 $\pm$ 0.06   \\
    MaPLe & ViT B16 & 8  & \checkmark & 1 & 0.1 & 69.66 $\pm$ 0.16 & 57.60 $\pm$ 0.81  \\
    MaPLe & ViT B16 & 4  & \checkmark & 1 & 0.1 & 66.96 $\pm$ 0.31 & 53.61 $\pm$ 0.55  \\
    MaPLe & ViT B16 & 2  & \checkmark & 1 & 0.1 & 66.74 $\pm$ 0.36 & 53.54 $\pm$ 0.76  \\
    MaPLe & ViT B16 & 1  & \checkmark & 1 & 0.1 & 63.46 $\pm$ 0.14 & 50.49 $\pm$ 1.01  \\
    \hline
    \end{tabular}
    }
    \caption{Performance of few-shot MaPLe on SUN. Ablations cover different NCL strengths $\lambda$.}
    \label{tab:full_sun_maple}
\end{table}

\begin{table}[H]
    \centering
    \adjustbox{max width=0.9\linewidth}{
    \begin{tabular}{|cccccc|c|}
    \hline
    Method  & Encoder & K-shot & w/ ProTeCt & $\lambda$ & $\beta$  & MTA \\
    \hline
    \hline
    CoOp & ViT B16 & 16 &  & N/A & N/A & 52.99 \\
    CoOp & ViT B16 & 8  &  & N/A & N/A & 55.24 \\
    CoOp & ViT B16 & 4  &  & N/A & N/A & 49.48 \\
    CoOp & ViT B16 & 2  &  & N/A & N/A & 51.94 \\
    CoOp & ViT B16 & 1  &  & N/A & N/A & 51.20 \\
    \hline
    CoOp & ViT B16 & 16 & \checkmark & 0.5 & 0.1 & 83.51 \\
    CoOp & ViT B16 & 8  & \checkmark & 0.5 & 0.1 & 81.34 \\
    CoOp & ViT B16 & 4  & \checkmark & 0.5 & 0.1 & 80.30 \\
    CoOp & ViT B16 & 2  & \checkmark & 0.5 & 0.1 & 76.59 \\
    CoOp & ViT B16 & 1  & \checkmark & 0.5 & 0.1 & 76.25 \\
    \hline
    \hline
    MaPLe & ViT B16 & 16 &  & N/A & N/A & 54.29 \\
    MaPLe & ViT B16 & 8  &  & N/A & N/A & 53.24 \\
    MaPLe & ViT B16 & 4  &  & N/A & N/A & 55.79 \\
    MaPLe & ViT B16 & 2  &  & N/A & N/A & 51.30 \\
    MaPLe & ViT B16 & 1  &  & N/A & N/A & 50.31 \\
    \hline
    MaPLe & ViT B16 & 16 & \checkmark & 0.5 & 0.1 & 82.27 \\
    MaPLe & ViT B16 & 8  & \checkmark & 0.5 & 0.1 & 80.71 \\
    MaPLe & ViT B16 & 4  & \checkmark & 0.5 & 0.1 & 79.10 \\
    MaPLe & ViT B16 & 2  & \checkmark & 0.5 & 0.1 & 77.55 \\
    MaPLe & ViT B16 & 1  & \checkmark & 0.5 & 0.1 & 76.73 \\
    \hline
    \end{tabular}
    }
    \caption{Performance of MTA for both few-shot CoOp and MaPLe on Sun.}
    \label{tab:full_sun_treecut}
\end{table}

\section{Complete Table of ImageNet Experiments}\label{sec:full_imagenet}
In this section, we report the complete experiment result conducted on ImageNet. Table~\ref{tab:full_imagenet_clip_cocoop} first show the performance of CLIP and CoCoOp as a complement of Table 1 in the main paper. Note that none of the CLIP features (e.g. ViT B32, ViT B16, RN50, RN101) nor existing prompt tuning methods help the HCA metric. 
Table~\ref{tab:full_imagenet_coop} and Table~\ref{tab:full_imagenet_maple} show the results of vanilla CoOp and MaPLe, and their results after adding ProTeCt. Table~\ref{tab:full_imagenet_treecut} further compares the MTA result of vanilla CoOp and MaPLe, and their ProTeCt counterpart. Clearly, adding ProTeCt boosts both HCA and MTA. Furthermore, we apply the model trained on ImageNet to its four variants. Table~\ref{tab:imagenetv2_domain}, Table~\ref{tab:imagenet_sketch_domain}, Table~\ref{tab:imagenet_a_domain} and  Table~\ref{tab:imagenet_r_domain} report the domain generalization results on ImageNetV2, ImageNet-sketch, ImageNet-A and ImageNet-R datasets for $Acc_{leaf}$, HCA and MTA. All four tables show that ProTeCt can not only improves the hierarchical consistency on the seen dataset, but also unseen datasets from other image domains.

\begin{table}[H]
    \centering
    \adjustbox{max width=0.9\linewidth}{
    \begin{tabular}{|cccccc|cc|}
    \hline
    Method  & Encoder & K-shot & w/ ProTeCt & $\lambda$ & $\beta$ & Leaf Acc. & HCA \\
    \hline
    \hline
    CLIP & ViT-B32 & 0 &  & N/A & N/A  & 63.31	 & 4.29   \\
    CLIP & ViT-B16 & 0  &  & N/A & N/A & 68.36   & 3.32 \\
    CLIP & RN50  & 0  &  & N/A & N/A   & 59.81	 & 4.16 \\
    CLIP & RN101 & 0  &  & N/A & N/A   & 62.30	 & 2.03 \\
    \hline
    CoCoOp & ViT-B16 & 16  &  & N/A & N/A   & 71.20 $\pm$ 0.13 & 2.92 $\pm$ 1.23 \\
    \hline
    \end{tabular}
    }
    \caption{Performance of zero-shot CLIP and 16-shot CoCoOp on ImageNet.}
    \label{tab:full_imagenet_clip_cocoop}
\end{table}

\begin{table}[H]
    \centering
    \adjustbox{max width=0.9\linewidth}{
    \begin{tabular}{|cccccc|cc|}
    \hline
    Method  & Encoder & K-shot & w/ ProTeCt & $\lambda$ & $\beta$ & Leaf Acc. & HCA \\
    \hline
    \hline
    CoOp & ViT B16 & 16 &  & N/A & N/A &  71.23 $\pm$ 0.67 & 2.99 $\pm$ 1.04    \\
    CoOp & ViT B16 & 8  &  & N/A & N/A &  69.40 $\pm$ 0.52 & 3.00 $\pm$ 0.58     \\
    CoOp & ViT B16 & 4  &  & N/A & N/A &  68.06 $\pm$ 0.42 & 2.95 $\pm$ 0.62     \\
    CoOp & ViT B16 & 2  &  & N/A & N/A &  65.46 $\pm$ 0.77 & 1.56 $\pm$ 0.17     \\
    CoOp & ViT B16 & 1  &  & N/A & N/A &  63.67 $\pm$ 0.85 & 1.59 $\pm$ 0.43      \\
    \hline
    CoOp & ViT B16 & 16 & \checkmark & 0 & 0.1 & 70.47 $\pm$ 0.22 & 27.81 $\pm$ 0.71  \\
    CoOp & ViT B16 & 8  & \checkmark & 0 & 0.1 & 70.03 $\pm$ 0.14 & 26.17 $\pm$ 0.52  \\
    CoOp & ViT B16 & 4  & \checkmark & 0 & 0.1 & 69.32 $\pm$ 0.11 & 21.99 $\pm$ 0.10  \\
    CoOp & ViT B16 & 2  & \checkmark & 0 & 0.1 & 68.09 $\pm$ 0.23 & 20.92 $\pm$ 1.02 \\
    CoOp & ViT B16 & 1  & \checkmark & 0 & 0.1 & 67.26 $\pm$ 0.65 & 18.69 $\pm$ 1.12  \\
    \hline 
    CoOp & ViT B16 & 8  & \checkmark & 0.5 & 0.1 & 70.27 $\pm$ 0.36 & 34.63 $\pm$ 0.33 \\
    CoOp & ViT B16 & 4  & \checkmark & 0.5 & 0.1 & 69.65 $\pm$ 0.41 & 31.84 $\pm$ 0.35 \\
    CoOp & ViT B16 & 2  & \checkmark & 0.5 & 0.1 & 68.09 $\pm$ 0.16 & 27.05 $\pm$ 0.27 \\
    CoOp & ViT B16 & 16 & \checkmark & 0.5 & 0.1 & 67.24 $\pm$ 0.24 & 26.09 $\pm$ 0.53 \\
    CoOp & ViT B16 & 1  & \checkmark & 0.5 & 0.1 & 66.69 $\pm$ 0.15 & 23.79 $\pm$ 0.15 \\
    \hline
    CoOp & ViT B16 & 16 & \checkmark & 1 & 0.1 & 69.92 $\pm$ 0.21 & 37.74 $\pm$ 0.12 \\
    CoOp & ViT B16 & 8  & \checkmark & 1 & 0.1 & 69.34 $\pm$ 0.17 & 34.66 $\pm$ 0.55 \\
    CoOp & ViT B16 & 4  & \checkmark & 1 & 0.1 & 68.06 $\pm$ 0.44 & 30.87 $\pm$ 0.32 \\
    CoOp & ViT B16 & 2  & \checkmark & 1 & 0.1 & 67.12 $\pm$ 0.35 & 26.34 $\pm$ 1.1 \\
    CoOp & ViT B16 & 1  & \checkmark & 1 & 0.1 & 66.11 $\pm$ 0.50 & 25.79 $\pm$ 0.06 \\
    \hline
    \end{tabular}
    }
    \caption{Performance of few-shot CoOp on ImageNet. Ablations cover different NCL strengths $\lambda$.}
    \label{tab:full_imagenet_coop}
\end{table}

\begin{table}[H]
    \centering
    \adjustbox{max width=0.9\linewidth}{
    \begin{tabular}{|cccccc|cc|}
    \hline
    Method  & Encoder & K-shot & w/ ProTeCt & $\lambda$ & $\beta$  & Leaf Acc. & HCA \\
    \hline
    \hline
    MaPLe & ViT B16 & 16 &  & N/A & N/A & 70.70	$\pm$ 0.11 & 4.15 $\pm$ 1.05   \\
    MaPLe & ViT B16 & 8  &  & N/A & N/A & 70.44	$\pm$ 0.06 & 4.32 $\pm$ 0.90 \\
    MaPLe & ViT B16 & 4  &  & N/A & N/A & 70.20	$\pm$ 0.06 & 2.95 $\pm$ 0.87 \\
    MaPLe & ViT B16 & 2  &  & N/A & N/A & 69.74	$\pm$ 0.25 & 4.27 $\pm$ 1.32 \\
    MaPLe & ViT B16 & 1  &  & N/A & N/A & 68.91	$\pm$ 0.13 & 2.97 $\pm$ 1.08 \\
    \hline
    MaPLe & ViT B16 & 16 & \checkmark & 0 & 0.1 &  70.08 $\pm$ 0.26 & 23.38 $\pm$ 1.43   \\
    MaPLe & ViT B16 & 8  & \checkmark & 0 & 0.1 &  69.00 $\pm$ 0.26 & 21.71 $\pm$ 0.64 \\
    MaPLe & ViT B16 & 4  & \checkmark & 0 & 0.1 &  68.50 $\pm$ 0.41 & 19.03 $\pm$ 0.21 \\
    MaPLe & ViT B16 & 2  & \checkmark & 0 & 0.1 &  67.45 $\pm$ 0.32 & 17.54 $\pm$ 0.52 \\
    MaPLe & ViT B16 & 1  & \checkmark & 0 & 0.1 &  67.03 $\pm$ 0.11 & 16.54 $\pm$ 0.32 \\
    \hline
    MaPLe & ViT B16 & 16 & \checkmark & 0.5 & 0.1 & 69.59 $\pm$ 0.25 & 27.74 $\pm$ 1.31   \\
    MaPLe & ViT B16 & 8  & \checkmark & 0.5 & 0.1 & 69.06 $\pm$ 0.49 & 25.25 $\pm$ 0.52   \\
    MaPLe & ViT B16 & 4  & \checkmark & 0.5 & 0.1 & 68.13 $\pm$ 0.01 & 25.25 $\pm$ 0.12   \\
    MaPLe & ViT B16 & 2  & \checkmark & 0.5 & 0.1 & 67.45 $\pm$ 0.43 & 20.14 $\pm$ 1.07   \\
    MaPLe & ViT B16 & 1  & \checkmark & 0.5 & 0.1 & 66.80 $\pm$ 0.26 & 20.62 $\pm$ 0.65   \\
    \hline
    MaPLe & ViT B16 & 16 & \checkmark & 1 & 0.1 &  69.52 $\pm$ 0.71 & 31.24 $\pm$ 1.02    \\
    MaPLe & ViT B16 & 8  & \checkmark & 1 & 0.1 &  68.48 $\pm$ 0.06 & 26.92 $\pm$ 0.42   \\
    MaPLe & ViT B16 & 4  & \checkmark & 1 & 0.1 &  68.59 $\pm$ 0.17 & 26.28 $\pm$ 0.31   \\
    MaPLe & ViT B16 & 2  & \checkmark & 1 & 0.1 &  67.12 $\pm$ 0.11 & 22.96 $\pm$ 0.05   \\
    MaPLe & ViT B16 & 1  & \checkmark & 1 & 0.1 &  66.16 $\pm$ 0.88 & 20.44 $\pm$ 0.77   \\
    \hline
    \end{tabular}
    }
    \caption{Performance of few-shot MaPLe on ImageNet. Ablations cover different NCL strengths $\lambda$.}
    \label{tab:full_imagenet_maple}
\end{table}

\begin{table}[H]
    \centering
    \adjustbox{max width=0.9\linewidth}{
    \begin{tabular}{|cccccc|c|}
    \hline
    Method  & Encoder & K-shot & w/ ProTeCt & $\lambda$ & $\beta$  & MTA \\
    \hline
    \hline
    CoOp & ViT B16 & 16 &  & N/A & N/A & 46.98 \\
    CoOp & ViT B16 & 8  &  & N/A & N/A & 46.04 \\
    CoOp & ViT B16 & 4  &  & N/A & N/A & 42.57 \\
    CoOp & ViT B16 & 2  &  & N/A & N/A & 44.89 \\
    CoOp & ViT B16 & 1  &  & N/A & N/A & 40.52 \\
    \hline
    CoOp & ViT B16 & 16 & \checkmark & 0.5 & 0.1 & 88.61 \\
    CoOp & ViT B16 & 8  & \checkmark & 0.5 & 0.1 & 87.86 \\
    CoOp & ViT B16 & 4  & \checkmark & 0.5 & 0.1 & 87.37 \\
    CoOp & ViT B16 & 2  & \checkmark & 0.5 & 0.1 & 86.14 \\
    CoOp & ViT B16 & 1  & \checkmark & 0.5 & 0.1 & 86.14 \\
    \hline
    \hline
    MaPLe & ViT B16 & 16 &  & N/A & N/A & 48.29 \\
    MaPLe & ViT B16 & 8  &  & N/A & N/A & 45.84 \\
    MaPLe & ViT B16 & 4  &  & N/A & N/A & 51.84 \\
    MaPLe & ViT B16 & 2  &  & N/A & N/A & 48.17 \\
    MaPLe & ViT B16 & 1  &  & N/A & N/A & 48.16 \\
    \hline
    MaPLe & ViT B16 & 16 & \checkmark & 0.5 & 0.1 & 87.87 \\
    MaPLe & ViT B16 & 8  & \checkmark & 0.5 & 0.1 & 87.26 \\
    MaPLe & ViT B16 & 4  & \checkmark & 0.5 & 0.1 & 86.85 \\
    MaPLe & ViT B16 & 2  & \checkmark & 0.5 & 0.1 & 85.93 \\
    MaPLe & ViT B16 & 1  & \checkmark & 0.5 & 0.1 & 85.18 \\
    \hline
    \end{tabular}
    }
    \caption{Performance of MTA for both few-shot CoOp and MaPLe on ImageNet.}
    \label{tab:full_imagenet_treecut}
\end{table}

\begin{table}[H]
    \centering
    \adjustbox{max width=0.9\linewidth}{
    \begin{tabular}{|cccccc|ccc|}
    \hline
    Method  & Encoder & K-shot & w/ ProTeCt & $\lambda$ & $\beta$ & Leaf Acc. & HCA  & MTA \\
    \hline
    \hline
    CoOp & ViT B16 & 16  &          & N/A & N/A   & 64.01 & 2.31 & 43.74  \\
    CoOp & ViT B16 & 8   &          & N/A & N/A   & 62.20 & 2.62 & 43.30  \\
    CoOp & ViT B16 & 4   &          & N/A & N/A   & 61.51 & 2.48 & 40.68  \\
    CoOp & ViT B16 & 2   &          & N/A & N/A   & 58.68 & 1.35 & 42.84  \\
    CoOp & ViT B16 & 1   &          & N/A & N/A   & 56.43 & 1.51 & 38.27  \\
    \hline
    CoOp & ViT B16 & 16  & \checkmark & 1 & 0.1   & 62.60 & 32.84 & 86.66  \\
    CoOp & ViT B16 & 8   & \checkmark & 1 & 0.1   & 62.15 & 30.65 & 85.84  \\
    CoOp & ViT B16 & 4   & \checkmark & 1 & 0.1   & 61.24 & 26.85 & 85.52  \\
    CoOp & ViT B16 & 2   & \checkmark & 1 & 0.1   & 60.42 & 23.22 & 84.38  \\
    CoOp & ViT B16 & 1   & \checkmark & 1 & 0.1   & 60.16 & 22.95 & 84.38  \\
    \hline
    \hline
    MaPLe & ViT B16 & 16  &          & N/A & N/A   & 64.15	& 1.97 & 45.93  \\
    MaPLe & ViT B16 & 8   &          & N/A & N/A   & 62.76	& 1.99 & 43.98  \\
    MaPLe & ViT B16 & 4   &          & N/A & N/A   & 63.45	& 2.51 & 49.41  \\
    MaPLe & ViT B16 & 2   &          & N/A & N/A   & 61.75	& 2.81 & 45.92  \\
    MaPLe & ViT B16 & 1   &          & N/A & N/A   & 61.78	& 2.18 & 45.50  \\
    \hline
    MaPLe & ViT B16 & 16  & \checkmark & 1 & 0.1   & 62.77 & 27.86 & 86.14  \\
    MaPLe & ViT B16 & 8   & \checkmark & 1 & 0.1   & 61.42 & 23.45 & 85.51  \\
    MaPLe & ViT B16 & 4   & \checkmark & 1 & 0.1   & 61.89 & 22.92 & 85.17  \\
    MaPLe & ViT B16 & 2   & \checkmark & 1 & 0.1   & 60.43 & 20.10 & 84.23  \\
    MaPLe & ViT B16 & 1   & \checkmark & 1 & 0.1   & 59.14 & 17.89 & 83.27  \\
    \hline
    \end{tabular}
    }
    \caption{Domain generalization on ImageNetv2 dataset using CoOp and MaPLe.}
    \label{tab:imagenetv2_domain}
\end{table}

\begin{table}[H]
    \centering
    \adjustbox{max width=0.9\linewidth}{
    \begin{tabular}{|cccccc|ccc|}
    \hline
    Method  & Encoder & K-shot & w/ ProTeCt & $\lambda$ & $\beta$  & Leaf Acc. & HCA & MTA\\
    \hline
    \hline
    CoOp & ViT B16 & 16  &          & N/A & N/A   & 47.82 & 1.39  & 38.58 \\
    CoOp & ViT B16 & 8   &          & N/A & N/A   & 45.93 & 2.10  & 42.56 \\
    CoOp & ViT B16 & 4   &          & N/A & N/A   & 44.60 & 1.41  & 36.52 \\
    CoOp & ViT B16 & 2   &          & N/A & N/A   & 42.17 & 0.96  & 36.01 \\
    CoOp & ViT B16 & 1   &          & N/A & N/A   & 41.38 & 1.11  & 33.61 \\
    \hline
    CoOp & ViT B16 & 16  & \checkmark & 1 & 0.1   & 46.80 & 20.73  & 82.60 \\
    CoOp & ViT B16 & 8   & \checkmark & 1 & 0.1   & 46.91 & 19.71  & 82.11 \\
    CoOp & ViT B16 & 4   & \checkmark & 1 & 0.1   & 46.53 & 17.69  & 82.07 \\
    CoOp & ViT B16 & 2   & \checkmark & 1 & 0.1   & 45.40 & 15.49  & 80.82 \\
    CoOp & ViT B16 & 1   & \checkmark & 1 & 0.1   & 44.75 & 13.88  & 80.64 \\
    \hline
    \hline
    MaPLe & ViT B16 & 16  &          & N/A & N/A   & 48.97 & 1.58 & 43.37  \\
    MaPLe & ViT B16 & 8   &          & N/A & N/A   & 47.55 & 1.66 & 45.26  \\
    MaPLe & ViT B16 & 4   &          & N/A & N/A   & 48.20 & 2.45 & 53.31  \\
    MaPLe & ViT B16 & 2   &          & N/A & N/A   & 46.86 & 1.01 & 42.55  \\
    MaPLe & ViT B16 & 1   &          & N/A & N/A   & 46.79 & 1.70 & 45.26  \\
    \hline
    MaPLe & ViT B16 & 16  & \checkmark & 1 & 0.1   & 47.47 & 17.77  & 82.52 \\
    MaPLe & ViT B16 & 8   & \checkmark & 1 & 0.1   & 46.60 & 15.31  & 82.04 \\
    MaPLe & ViT B16 & 4   & \checkmark & 1 & 0.1   & 47.23 & 14.95  & 81.67 \\
    MaPLe & ViT B16 & 2   & \checkmark & 1 & 0.1   & 45.95 & 13.32  & 80.87 \\
    MaPLe & ViT B16 & 1   & \checkmark & 1 & 0.1   & 44.92 & 11.24  & 79.94 \\
    \hline
    \end{tabular}
    }
    \caption{Domain generalization on ImageNet-sketch dataset using CoOp and MaPLe.}
    \label{tab:imagenet_sketch_domain}
\end{table}

\begin{table}[H]
    \centering
    \adjustbox{max width=0.9\linewidth}{
    \begin{tabular}{|cccccc|ccc|}
    \hline
    Method  & Encoder & K-shot & w/ ProTeCt & $\lambda$ & $\beta$  & Leaf Acc. & HCA & MTA \\
    \hline
    \hline
    CoOp & ViT B16 & 16  &          & N/A & N/A   & 50.28 & 2.97 & 52.56  \\
    CoOp & ViT B16 & 8   &          & N/A & N/A   & 48.08 & 4.19 & 45.05  \\
    CoOp & ViT B16 & 4   &          & N/A & N/A   & 48.43 & 2.97 & 41.20  \\
    CoOp & ViT B16 & 2   &          & N/A & N/A   & 46.56 & 1.95 & 52.47  \\
    CoOp & ViT B16 & 1   &          & N/A & N/A   & 45.92 & 1.76 & 47.54  \\
    \hline
    CoOp & ViT B16 & 16  & \checkmark & 1 & 0.1   & 49.08 & 22.45 & 78.21  \\
    CoOp & ViT B16 & 8   & \checkmark & 1 & 0.1   & 49.29 & 24.00 & 79.47  \\
    CoOp & ViT B16 & 4   & \checkmark & 1 & 0.1   & 48.39 & 18.11 & 76.95  \\
    CoOp & ViT B16 & 2   & \checkmark & 1 & 0.1   & 48.81 & 20.00 & 78.11  \\
    CoOp & ViT B16 & 1   & \checkmark & 1 & 0.1   & 48.95 & 20.52 & 76.95  \\
    \hline
    \hline
    MaPLe & ViT B16 & 16  &          & N/A & N/A   & 50.61 & 2.31  & 54.88 \\
    MaPLe & ViT B16 & 8   &          & N/A & N/A   & 48.41 & 5.31  & 55.97 \\
    MaPLe & ViT B16 & 4   &          & N/A & N/A   & 50.23 & 4.95  & 57.07 \\
    MaPLe & ViT B16 & 2   &          & N/A & N/A   & 48.49 & 9.80  & 59.90 \\
    MaPLe & ViT B16 & 1   &          & N/A & N/A   & 47.55 & 3.52  & 55.48 \\
    \hline
    MaPLe & ViT B16 & 16  & \checkmark & 1 & 0.1   & 47.41 & 19.75 & 77.46  \\
    MaPLe & ViT B16 & 8   & \checkmark & 1 & 0.1   & 46.15 & 16.49 & 75.88  \\
    MaPLe & ViT B16 & 4   & \checkmark & 1 & 0.1   & 47.35 & 17.39 & 77.64  \\
    MaPLe & ViT B16 & 2   & \checkmark & 1 & 0.1   & 49.15 & 16.23 & 77.71  \\
    MaPLe & ViT B16 & 1   & \checkmark & 1 & 0.1   & 47.15 & 16.03 & 76.81  \\
    \hline
    \end{tabular}
    }
    \caption{Domain generalization on ImageNet-A dataset using CoOp and MaPLe.}
    \label{tab:imagenet_a_domain}
\end{table}

\begin{table}[H]
    \centering
    \adjustbox{max width=0.9\linewidth}{
    \begin{tabular}{|cccccc|ccc|}
    \hline
    Method  & Encoder & K-shot & w/ ProTeCt & $\lambda$ & $\beta$  & Leaf Acc. & HCA & MTA\\
    \hline
    \hline
    CoOp & ViT B16 & 16  &          & N/A & N/A   & 75.83 & 18.49  & 64.13 \\
    CoOp & ViT B16 & 8   &          & N/A & N/A   & 74.79 & 5.91   & 49.56 \\
    CoOp & ViT B16 & 4   &          & N/A & N/A   & 73.99 & 14.85  & 61.40  \\
    CoOp & ViT B16 & 2   &          & N/A & N/A   & 70.94 & 16.32  & 56.67  \\
    CoOp & ViT B16 & 1   &          & N/A & N/A   & 69.84 & 11.74  & 55.31  \\
    \hline
    CoOp & ViT B16 & 16  & \checkmark & 1 & 0.1   & 74.94 & 31.18 & 75.59  \\
    CoOp & ViT B16 & 8   & \checkmark & 1 & 0.1   & 75.51 & 37.96 & 81.11  \\
    CoOp & ViT B16 & 4   & \checkmark & 1 & 0.1   & 74.23 & 29.69 & 75.54  \\
    CoOp & ViT B16 & 2   & \checkmark & 1 & 0.1   & 74.86 & 28.67 & 78.17  \\
    CoOp & ViT B16 & 1   & \checkmark & 1 & 0.1   & 74.26 & 27.46 & 76.48  \\
    \hline
    \hline
    MaPLe & ViT B16 & 16  &          & N/A & N/A   & 76.61 & 20.67  & 63.06 \\
    MaPLe & ViT B16 & 8   &          & N/A & N/A   & 76.48 & 18.92  & 67.30 \\
    MaPLe & ViT B16 & 4   &          & N/A & N/A   & 76.83 & 21.06  & 64.30 \\
    MaPLe & ViT B16 & 2   &          & N/A & N/A   & 75.85 & 19.84  & 60.86 \\
    MaPLe & ViT B16 & 1   &          & N/A & N/A   & 74.55 & 18.85  & 62.48 \\
    \hline
    MaPLe & ViT B16 & 16  & \checkmark & 1 & 0.1   & 75.70 & 32.58 & 77.99  \\
    MaPLe & ViT B16 & 8   & \checkmark & 1 & 0.1   & 75.98 & 30.97 & 77.57  \\
    MaPLe & ViT B16 & 4   & \checkmark & 1 & 0.1   & 76.31 & 29.28 & 78.52  \\
    MaPLe & ViT B16 & 2   & \checkmark & 1 & 0.1   & 75.01 & 23.94 & 72.73  \\
    MaPLe & ViT B16 & 1   & \checkmark & 1 & 0.1   & 74.60 & 25.20 & 75.72  \\
    \hline
    \end{tabular}
    }
    \caption{Domain generalization on ImageNet-R dataset using CoOp and MaPLe.}
    \label{tab:imagenet_r_domain}
\end{table}

\section{Additional Taxonomies}\label{sec:addtional}
To investigate the robustness of ProTeCt across hierarchies, we consider the FGVC Aircraft~\cite{Maji2013FineGrainedVC} dataset and the RSI-CB~\cite{li2020RSI-CB} satellite dataset. These datasets have their built-in hierarchies, which beyond differing from those of SUN~\cite{sun397} and WordNet~\cite{WordNet}, are a technical hierarchy of fine-grained aircraft classes and satellite image classes, respectively. Table~\ref{tab:coop_aircraft} and Table~\ref{tab:rsi-cb} summarize the CoOp results for these experiments, showing that ProTeCt improves performance under all metrics. This illustrates its taxonomy robustness.

\begin{table}[H]
    \centering
    \resizebox{0.9\linewidth}{!}{
    \setlength{\tabcolsep}{8pt}
    \begin{tabular}{|cc|ccc|}
    \hline
    K-shot & w/ ProTeCt   & $Acc_{leaf}$ & HCA & MTA (25) \\
    \hline
    \hline
     16  &  & 41.88 & 17.82 & 21.11\\
     16 & \checkmark & 42.00 & 29.94 & 32.95\\
     & &  {\color{red}(+0.12)} &  {\color{red}(+12.12)} &  {\color{red}(+11.84)}\\
    \hline
     \hline
      1 & & 23.61 & 11.55 & 16.77 \\
     1 & \checkmark & 27.30 & 16.47 & 24.67\\
     & &  {\color{red}(+3.69)} &  {\color{red}(+4.92)} &  {\color{red}(+7.90)}\\
    \hline    
    \end{tabular}
    }
    \caption{Comparison of CoOp with/without ProTeCt on FGVC Aircraft~\cite{Maji2013FineGrainedVC} dataset.}
    \label{tab:coop_aircraft}
\end{table}

\begin{table}[H]
    \centering
    \resizebox{0.9\linewidth}{!}{
    \setlength{\tabcolsep}{8pt}
    \begin{tabular}{|cc|ccc|}
    \hline
    K-shot & w/ ProTeCt   & $Acc_{leaf}$ & HCA & MTA (25) \\
    \hline
    \hline
     16  &  & 91.79 & 43.50 & 64.49\\
     16 & \checkmark & 93.21 & 85.21 & 91.44\\
     & &  {\color{red}(+1.42)} &  {\color{red}(+41.71)} &  {\color{red}(+26.95)}\\
    \hline
     \hline
      1 & & 63.93 & 32.29 & 52.17 \\
     1 & \checkmark & 65.00 & 48.36 & 67.05\\
     & &  {\color{red}(+1.07)} &  {\color{red}(+16.07)} &  {\color{red}(+14.88)}\\
    \hline    
    \end{tabular}
    }
    \caption{Comparison of CoOp with/without ProTeCt on RSI-CB~\cite{li2020RSI-CB} satellite dataset.}
    \label{tab:rsi-cb}
\end{table}

\section{Complete Ablation Results}\label{sec:full_ablade}
This section complements Figure 4(a) and 4(b) in the main paper with the error bar. Table~\ref{tab:lambda_ablade_cifar100_coop} and Table ~\ref{tab:td_ablade_cifar100_coop} show how the NCL strength  $\lambda$ and  tree dropout rates $\beta$ affect the $Acc_{leaf}$ and HCA. Please refer to Section 6.3 of the main paper for more discussion.

\begin{table}[H]
    \centering
    \adjustbox{max width=0.9\linewidth}{
    \begin{tabular}{|cccccc|cc|}
    \hline
    Method  & Encoder & K-shot & w/ ProTeCt & $\lambda$ & $\beta$  & Leaf Acc. & HCA \\
    \hline
    \hline
    CoOp & ViT B16 & 1  & \checkmark & 0   & 0.1   & 64.77 $\pm$ 1.37 & 32.93 $\pm$ 0.42   \\
    CoOp & ViT B16 & 1  & \checkmark & 0.1 & 0.1   & 66.18 $\pm$ 0.51 & 39.41 $\pm$ 0.78 \\
    CoOp & ViT B16 & 1  & \checkmark & 0.3 & 0.1   & 67.13 $\pm$ 0.66 & 40.82 $\pm$ 0.33 \\
    CoOp & ViT B16 & 1  & \checkmark & 0.5 & 0.1   & 66.88 $\pm$ 0.21 & 41.01 $\pm$ 1.18 \\
    CoOp & ViT B16 & 1  & \checkmark & 0.7 & 0.1   & 66.50 $\pm$ 0.71 & 40.39 $\pm$ 0.37 \\
    CoOp & ViT B16 & 1  & \checkmark & 1   & 0.1   & 63.84 $\pm$ 1.51 & 40.05 $\pm$ 1.48 \\
    \hline
    \end{tabular}
    }
    \caption{Ablation of different NCL strength  $\lambda$ on Cifar100 using CoOp 1 shot setting.}
    \label{tab:lambda_ablade_cifar100_coop}
\end{table}

\begin{table}[H]
    \centering
    \adjustbox{max width=0.9\linewidth}{
    \begin{tabular}{|cccccc|cc|}
    \hline
    Method  & Encoder & K-shot & w/ ProTeCt & $\lambda$ & $\beta$    & Leaf Acc. & HCA \\
    \hline
    \hline
    CoOp & ViT B16 & 16  & \checkmark & 1   & 0     & 70.86 $\pm$ 0.59 & 54.39 $\pm$ 0.68   \\
    CoOp & ViT B16 & 16  & \checkmark & 1   & 0.1   & 73.26 $\pm$ 0.66 & 58.01 $\pm$ 0.43 \\
    CoOp & ViT B16 & 16  & \checkmark & 1   & 0.2   & 72.48 $\pm$ 0.57 & 59.32 $\pm$ 0.21 \\
    CoOp & ViT B16 & 16  & \checkmark & 1   & 0.3   & 71.49 $\pm$ 0.36 & 58.82 $\pm$ 0.12 \\
    CoOp & ViT B16 & 16  & \checkmark & 1   & 0.4   & 70.15 $\pm$ 0.75 & 57.93 $\pm$ 0.50 \\
    CoOp & ViT B16 & 16  & \checkmark & 1   & 0.5   & 69.22 $\pm$ 0.32 & 56.66 $\pm$ 0.41 \\
    CoOp & ViT B16 & 16  & \checkmark & 1   & 0.6   & 68.35 $\pm$ 0.78 & 53.75 $\pm$ 0.85 \\
    CoOp & ViT B16 & 16  & \checkmark & 1   & 0.7   & 66.58 $\pm$ 0.45 & 53.27 $\pm$ 0.42 \\
    CoOp & ViT B16 & 16  & \checkmark & 1   & 0.8   & 66.62 $\pm$ 0.38 & 50.74 $\pm$ 1.05 \\
    CoOp & ViT B16 & 16  & \checkmark & 1   & 0.9   & 66.77 $\pm$ 0.38 & 49.76 $\pm$ 1.02 \\
    \hline
    \end{tabular}
    }
    \caption{Ablation of different tree dropout rates $\beta$ on Cifar100 using CoOp 16 shot setting.}
    \label{tab:td_ablade_cifar100_coop}
\end{table}

\section{Visualizations}\label{sec:vis}
\begin{figure}[b!]
    \centering
    \begin{minipage}[h]{0.49\linewidth}
        \includegraphics[width=\linewidth]{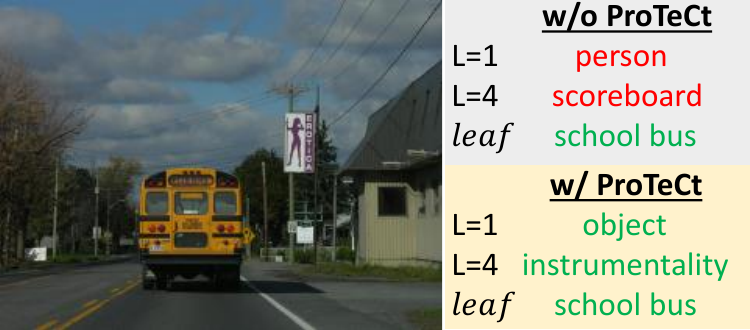}
        \includegraphics[width=\linewidth]{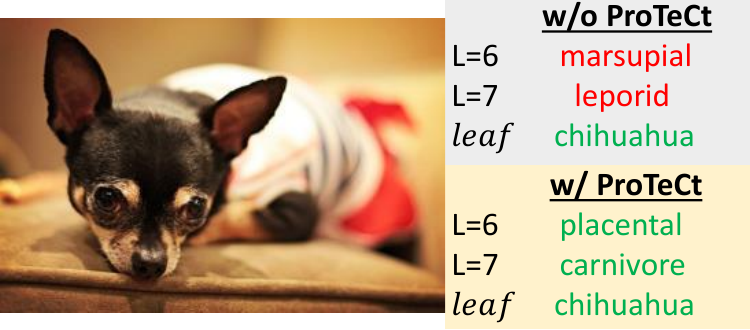}
    \end{minipage}
    \hfill
    \begin{minipage}[h]{0.49\linewidth}
        \includegraphics[width=\linewidth]{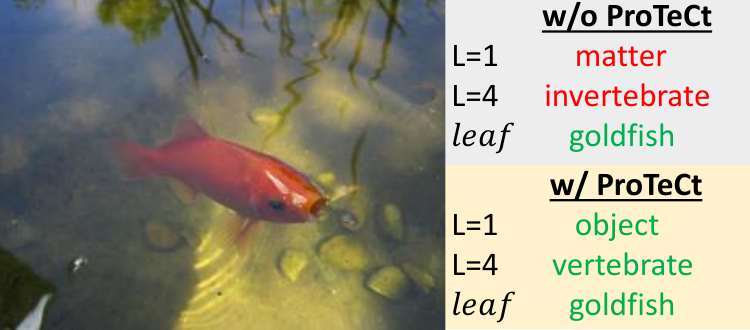}
        \includegraphics[width=\linewidth]{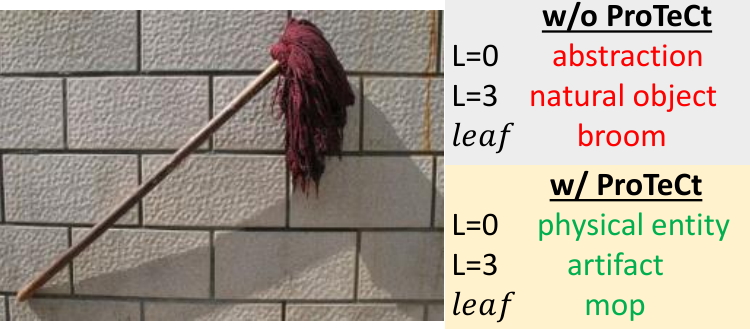}
    \end{minipage}
    \caption{ImageNet visual examples at multiple hierarchy levels. Correct/incorrect model predictions ({\color{Green} green}/{\color{red} red}) of CoOp w/ and w/o ProTeCt, respectively. L denotes the tree level.}
    \label{fig:multi-hier}
\end{figure}

\begin{figure}[t!]
    \centering
       \resizebox{\linewidth}{!}{
        \begin{tabular}{ccc}
        \setlength{\tabcolsep}{-1pt}
        \includegraphics[height=0.3\linewidth]{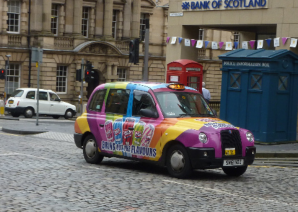} &
        \includegraphics[height=0.3\linewidth]{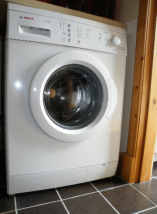} &
        \includegraphics[height=0.3\linewidth]{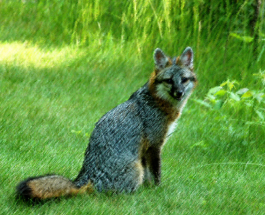} \\
        (a) & (b) & (c) \\
        \vspace{-7pt}
        \end{tabular}
        }
        \captionof{figure}{ImageNetv2 visual examples: (a): [{\color{blue}{Taxicab}}, {\color{red}{Teddy bear}}], (b): [{\color{blue}{Washing machine}}, {\color{red}{Bath towel}}], (c):[{\color{blue}{Grey fox}}, {\color{red}{Marsupial}}].}
        \label{fig:appendix_vis_imagenetv2}
        \vspace{10pt}
\end{figure}

\begin{figure}[t!]
    \centering
    \resizebox{\linewidth}{!}{
        \begin{tabular}{cccc}
        \setlength{\tabcolsep}{-1pt}
        \includegraphics[height=0.3\linewidth]{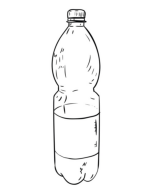} &
        \includegraphics[height=0.3\linewidth]{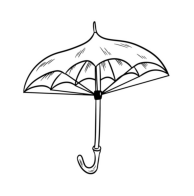} &
        \includegraphics[height=0.3\linewidth]{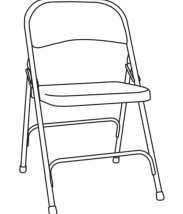} &
        \includegraphics[height=0.3\linewidth]{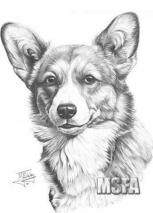} \\
        (a) & (b) & (c) & (d) \\
        \vspace{-7pt}
        \end{tabular}
        }
        \captionof{figure}{ImageNet-S visual examples: (a): [{\color{blue}{Water bottle}}, {\color{red}{Soap dispenser}}], (b): [{\color{blue}{Umbrella}}, {\color{red}{Lampshade}}], (c):[{\color{blue}{Folding chair}}, {\color{red}{Baby bed}}], (d):[{\color{blue}{Pembroke Welsh Corgi}}, {\color{red}{Marsupial}}].}
        \label{fig:appendix_vis_imagenets}
\end{figure}

\begin{figure}[t!]
    \centering
    \resizebox{\linewidth}{!}{
        \begin{tabular}{ccc}
        \setlength{\tabcolsep}{-1pt}
        \includegraphics[height=0.33\linewidth]{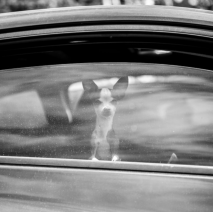} &
        \includegraphics[height=0.33\linewidth]{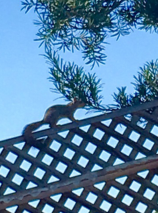} &
        \includegraphics[height=0.33\linewidth]{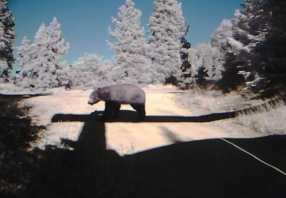} \\
        (a) & (b) & (c) \\
        \vspace{-7pt}
        \end{tabular}
    }
        \captionof{figure}{ImageNet-A visual examples: (a): [{\color{blue}{Chihuahua}}, {\color{red}{Cottontail rabbit}}], (b): [{\color{blue}{Fox squirrel}}, {\color{red}{Bird}}], (c):[{\color{blue}{American black bear}}, {\color{red}{Koala}}].}
        \label{fig:appendix_vis_imageneta}
\end{figure}

\begin{figure}[t!]
    \centering
    \resizebox{\linewidth}{!}{
        \begin{tabular}{ccc}
        \setlength{\tabcolsep}{-1pt}
        \includegraphics[height=0.3\linewidth]{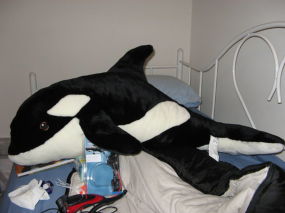} &
        \includegraphics[height=0.3\linewidth]{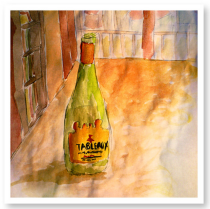} &
        \includegraphics[height=0.3\linewidth]{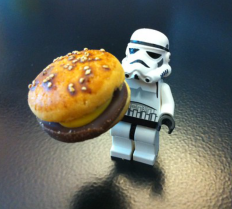} \\
        (a) & (b) & (c) \\
        \vspace{-7pt}
        \end{tabular}
    }
        \captionof{figure}{ImageNet-R visual examples: (a): [{\color{blue}{Killer whale}}, {\color{red}{Person}}], (b): [{\color{blue}{Wine bottle}}, {\color{red}{Fruit}}], (c):[{\color{blue}{Cheeseburger}}, {\color{red}{Ice cream}}].}
        \label{fig:appendix_vis_imagenetr}
\end{figure}

This section illustrates some misclassified examples of prior prompt tuning methods in ImageNet and its variants (i.e. ImageNetv2, ImageNet-S, ImageNet-A, ImageNet-R). Note that the hierarchy of these variants may differ from the one of ImageNet. The misclassification can occur in both coarse or fine-grained levels of the hierarchy. Note that ProTeCt can successfully classify all the illustrated examples at \textbf{every} hierarchy level in the examples shown in Figure~\ref{fig:multi-hier}-\ref{fig:appendix_vis_imagenetr}.
Figure~\ref{fig:multi-hier} presents the {\color{Green} correct}/{\color{red} incorrect} predictions of CoOp and its ProTeCt counterpart at multiple tree levels on ImageNet. CoOp~\cite{coop} fails to generate consistent predictions at different hierarchy levels, and even predicts incorrectly at coarser hierarchy levels when the predictions at the leaf level are correct.
More examples of the predictions on ImageNet variants are shown in Figure~\ref{fig:appendix_vis_imagenetv2}-\ref{fig:appendix_vis_imagenetr}, where [{\color{blue}{GT}}, {\color{red}{Prediction}}] shows the {\color{blue}{groundtruth}} and {\color{red}{incorrect prediction}} by vanilla prompt tuning.  

\end{document}